\RequirePackage{letltxmacro}
\LetLtxMacro{\LaTeXtextbf}{\textbf}

\documentclass{ieeeaccess}

\LetLtxMacro{\textbf}{\LaTeXtextbf}

\usepackage{cite}
\usepackage{amsmath,amssymb,amsfonts}
\usepackage{graphicx}
\usepackage{textcomp}
\usepackage[mathscr]{euscript}
\usepackage{comment}

\usepackage{hyperref}

\hypersetup{
    colorlinks=true,
    linkcolor=blue,
    filecolor=blue,      
    urlcolor=blue,
    citecolor=blue,
}

\usepackage{ragged2e}
\usepackage{amsthm}

\def\,{$\mskip\thinmuskip$} \def\!{$\mskip-\thinmuskip$}

\usepackage{algorithm}
\usepackage[algo2e]{algorithm2e} 
\usepackage{algorithmicx}
\usepackage{textgreek}
\usepackage{textcomp}
\newcommand{\var}{\texttt}

\def\BibTeX{{\rm B\kern-.05em{\sc i\kern-.025em b}\kern-.08em
    T\kern-.1667em\lower.7ex\hbox{E}\kern-.125emX}}

\begin{document}
\doi{-}

\title{ConvGeN: Convex space learning improves deep-generative oversampling for tabular imbalanced classification on smaller datasets}

\author{\uppercase{Kristian Schultz\authorrefmark{1}, Saptarshi Bej\authorrefmark{1}, Waldemar Hahn\authorrefmark{3}, Markus Wolfien\authorrefmark{3}, Prashant Srivastava\authorrefmark{1}, Olaf Wolkenhauer\authorrefmark{1,2}}}

\address[1]{Institute of Computer Science, University of Rostock, Germany}
\address[2]{Leibniz-Institute for Food Systems Biology, Technical University of Munich, Freising, 85354, Germany}
\address[3]{Institute for Medical Informatics and Biometry, Faculty of Medicine Carl Gustav Carus, Technische Universität Dresden, Germany}



\corresp{Corresponding author: Olaf Wolkenhauer (olaf.wolkenhauer@uni-rostock.de), ORCID: 0000-0001-6105-2937}

\begin{abstract}
\justify

Data is commonly stored in tabular format. Several fields of research are prone to small imbalanced tabular data. Supervised Machine Learning on such data is often difficult due to class imbalance. Synthetic data generation, i.e., oversampling, is a common remedy used to improve classifier performance. State-of-the-art linear interpolation approaches, such as LoRAS and ProWRAS can be used to generate synthetic samples from the convex space of the minority class to improve classifier performance in such cases. Deep generative networks are common deep learning approaches for synthetic sample generation, widely used for synthetic image generation.  However, their scope on synthetic tabular data generation in the context of imbalanced classification is not adequately explored. In this article, we show that existing deep generative models perform poorly compared to linear interpolation based approaches for imbalanced classification problems on smaller tabular datasets. To overcome this, we propose a deep generative model, \textit{ConvGeN} that combines the idea of convex space learning with deep generative models. ConvGeN learns the coefficients for the convex combinations of the minority class samples, such that the synthetic data is distinct enough from the majority class. Our benchmarking experiments demonstrate that our proposed model ConvGeN improves imbalanced classification on such small datasets, as compared to existing deep generative models, while being at-par with the existing linear interpolation approaches. Moreover, we discuss how our model can be used for synthetic tabular data generation in general, even outside the scope of data imbalance and thus, improves the overall applicability of convex space learning.
\end{abstract}


\titlepgskip=-15pt

\maketitle

\section{Introduction}\label{sec:introduction}

Tabular datasets are a popular and convenient format of storing data. Most data is stored in relational databases and are therefore tabular by nature. 
Yet, another major challenge associated to such fields is that often it is not possible to obtain balanced data for classification and decision-making problems. In fact, in many important practical problems, class imbalance intrinsically occurs. For example, in case of fraud detection problems it is natural that only a small fraction of the population would commit a fraud. It is well known that class imbalance makes it difficult for Machine Learning (ML) models to adapt well to the data and learn latent patterns because there are certain classes with only very few instances. Therefore, classification models tend to be biased towards the majority class \cite{SMOTE}. Classes with fewer number of instances are called \textit{minority classes}, while the larger class is called \textit{majority classes}. The region in the data manifold where these two classes intersect is often called the \textit{borderline region}, and is often of particular interest for ML applications. In this manuscript, we limit our investigation to binary classification problems and therefore address one majority and one minority class \cite{ProWRAS}.\par
One major approach of improving imbalanced classification is synthetic sample generation for the minority class to balance the classes in a dataset. Historically, the SMOTE algorithm proposed in 2002, has been the first algorithm to generate synthetic samples for the minority class \cite{SMOTE}. The algorithm is based on a linear interpolation-based approach, where a convex linear combination of two minority class samples that are closely located, is considered a synthetic sample. A convex combination is a linear combination of vectors, such that the sum of the coefficients of the linear combination is one \cite{ProWRAS}.
However, with time, researchers pointed out several limitations of the SMOTE algorithm, such as `over-generalization of the minority class' or that the samples generated in a minority data neighborhood has high variance \cite{LoRAS, ProWRAS, Blagus}. This can be interpreted as high variance of the synthetic samples in data neighborhoods, especially in the borderline region, that leads to undesirable interference of the synthetic samples with the majority class, and therefore, create confusion among the classifiers\cite{ProWRAS}. A recent article documents 85 SMOTE-based oversampling algorithms and performs a detailed benchmarking study on them, demonstrating that the problem of class imbalance is still open to research and improvement \cite{Comparison}.\par
Recently, the approach of \textit{convex space learning} has been proposed to model the convex space of the minority class in an improved manner to overcome the identified limitations problems of the SMOTE algorithms. The Localized Random Affine Shadowsampling (LoRAS) algorithm does this by generating synthetic samples using convex linear combinations of multiple shadowsamples, which are Gaussian noise added to the original minority samples in a minority data neighborhood \cite{LoRAS}. This approach reduces the variance of the synthetic samples and prevents them from interfering with the majority class \cite{LoRAS}. Another improved version of the LoRAS algorithm called ProWRAS makes the modeling of the convex space even more precise. The specialty of this algorithm is that, it offers four oversampling schemes depending on the choice of two of its parameters, enabling it to generate customized synthetic samples for several classifiers \cite{ProWRAS}. Oversampling algorithms before ProWRAS were known to perform well only for specific classifiers. However, the ProWRAS algorithm, unlike its predecessors, can be compatible with a wide range of classifiers \cite{ProWRAS}. We call this philosophy of rigorous modeling of the convex data space to sample synthetic data from it, and thereby enhance ML model performance for a specific task (in this case classification), as \textit{convex space learning} or \textit{convex space modeling}.\par 
\textit{Generative Adversarial Network (GAN)} is a very popular neural network-based model for synthetic data generation. Using two competing neural networks, \textit{generator} and \textit{discriminator}, GANs can create realistic synthetic data from random noise \cite{GAN}. While training the model, a random noise vector is passed into the generator, which creates synthetic samples from the noise, while the discriminator learns to differentiate between real samples against the synthetic samples generated by the generator. Over several iterations of competitive training, the generator learns to create more and more realistic synthetic data \cite{GAN}. A huge amount of research has been done on GANs with the goal of `realistic data generation'. Diverse models based on the GAN approach have been developed since the introduction of GANs in 2014 \cite{ProGAN, CycleGAN, StyleGAN}.\par
Although mostly such models are focused on synthetic image generation, since 2017, multiple deep-generative models have been developed focusing on synthetic data generation for tabular datasets. Note that, realistic data generation', while easy to perceive for image based data, is harder to realize or judge for tabular datasets, as there is no direct visual aid for such data. MedGAN \cite{MedGAN}, the first of such architectures, can handle either binary or count data. Another model TableGAN \cite{TableGAN}, based on deep convolutional GAN (DCGAN), also consisting of an additional classifier neural network can generate numerical and categorical values. TGAN or tabular GAN\cite{TGAN} handles multi-modal data in continuous variables through Gaussian Mixture Models (GMM) and can create categorical values with the help of Gumbel-Softmax as activation function, using a Long-Short-Term-Memory (LSTM) generator. Another model from 2019, CTGAN introduces conditional vectors for categorical values \cite{{CtGAN}}. CTAB-GAN proposed in 2021, is a combination of TableGAN and CTGAN \cite{CtabGAN}.\par 
However, GAN-based synthetic data generation has several natural limitations when it comes to smaller tabular datasets that are meant for imbalanced classification:
\begin{itemize}
    \item Generative adversarial models are not specifically aimed at generating synthetic data for classification tasks. If realistic data generation means the ability of a network to generate exact copies of the existing data, then intuitively, that does not add much additional information to classifiers. Clearly, for imbalanced classification, the variation in the synthetic data is a key factor and generating exact copies of the original data reduces it drastically.
    \item GANs are more difficult to train on small amounts of data \cite{limited}.
\end{itemize}
In this article, we will investigate through our experiments the effectiveness of oversampling using GAN-based models to improve classification on smaller tabular imbalanced data. Note that, we considered an empirical assumption that small imbalanced data means at most $150$ minority data points being used for synthetic data generation, while training a data rebalancing model and datasets containing at most $3500$ data points in total, with an imbalance ratio of at least $15:1$ (the ratio of the number of majority points to the number of minority points). We identified $14$ benchmarking datasets in the public domain satisfying our criterion of being `small' that are subsequently used for our benchmarking studies.\par
Moreover, we here propose a novel deep-generative model called \textit{Convex-space Generative Network} (ConvGeN) that combines the philosophy of linear interpolation based approaches with that of deep-generative networks to improve synthetic sample generation in the context of imbalanced classification for small datasets (as per our empirical assumption). The model consists of a generator-discriminator architecture. However, instead of competing among themselves as seen in regular GANs, in ConvGeN, they cooperate with each other. The generator takes randomly shuffled minority neighborhood batches as input, instead of taking noise as input in traditional models. The generator then creates convex linear combinations of the input neighborhood as synthetic samples. The discriminator learns to differentiate the synthetic samples from a shuffled batch of majority class samples (that can be chosen such that the majority batches are proximal to the respective minority batches) chosen as inputs at each training step. In this way, during the training process, the generator learns the coefficients of the convex linear combination that are more suitable for synthetic sample generation, in the sense that the synthetic samples can be identified distinctly by the discriminator. An additional bonus is that during the training, the discriminator is automatically trained as a classifier customized to the synthetic samples. \par 
Interestingly, we have chosen a `Repeater' model as a baseline of our experiments. This model creates exact copies of minority class samples to balance the two classes. We noticed that for smaller tabular datasets, state-of-the-art generative models specialized for tabular datasets like CTGAN and CTAB-GAN, in most cases have poor performance than the baseline. Our experiments show that ConvGeN performs at-par with the state-of-the-art interpolation based approach ProWRAS, while being superior to the traditional GAN, including CTGAN and CTAB-GAN, and in terms of popular metrics, such as F1-score and $\kappa$-score for imbalanced classification problems on tabular datasets of small size.\par
The novelty of ConvGeN is that it generates synthetic samples from the convex space, but unlike existing liner interpolation based models, such as SMOTE, LoRAS or ProWRAS, the coefficients of convex combinations are not chosen randomly or decided as a part of the algorithm, but are learned by the model itself during the training. The model thus provides a non-linearly linear' method for synthetic data generation, in the sense that the synthetic samples are convex linear combinations of the original samples, but the coefficients of the convex linear combinations are learned in a non-linear way.

\begin{figure*}[!htbp]

\centering
\includegraphics[width=.95\textwidth]{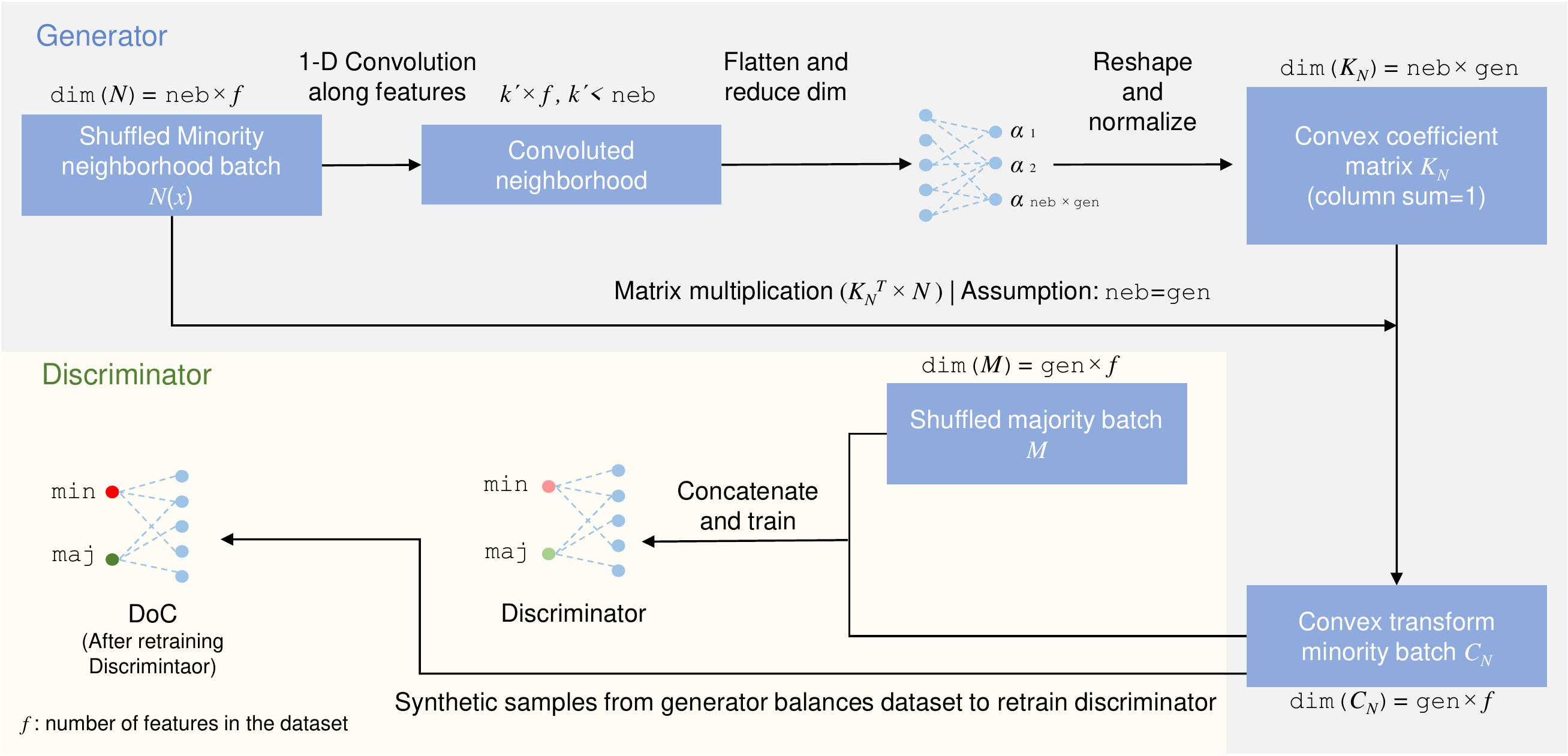} 
\caption{Figure showing the architecture of the ConvGeN model. The generator receives shuffled minority neighborhood batches as input and generates a convex combination of the minority neighborhood samples as output. The discriminator classifies these generated samples against a random majority batch of the same size as the minority neighborhood. The discriminator can be retrained as the DoC classifier after the training of the ConvGeN model.}
\label{fig:ConvGeN}
\end{figure*}

\section{Related research} \label{sec:Algo}
We have used six algorithms in total for synthetic data generation in our experiments. These algorithms also majorly consist of  related research for ConvGeN. Thus, we hereby discuss these related algorithms. A more detailed rationale of why these algorithms are closely related to ConvGeN is given in Section \ref{Protocols for benchmarking}. 
\subsection{Repeater}\label{sec:Algo:Repeater}
This model was used as a baseline for our experiments. The repeater model simply generates multiple copies of minority class samples to balance both the classes. Exact copies of the data points in the minority class are generated sequentially until the classes are balanced. Conventionally, the scenario with no oversampling is used as a benchmark \cite{LoRAS,ProWRAS}. However, since it is well-known that balanced datasets produce better classification performances, this is the simplest model one could build to balance the datasets. 

\subsection{ProWRAS}\label{sec:Algo:ProWRAS}
ProWRAS\cite{ProWRAS} (GIT link: \url{https://github.com/COSPOV/ProWRAS.git}) is a multi-schematic oversampling algorithm using convex linear combinations of minority shadowsamples (Gaussian noise added to original minority samples), for synthetic sample generation. SMOTE and its extension algorithms are known to perform with different efficiency for different classifiers. For example, although the LoRAS algorithm works very well for kNN, it produces relatively poor performance with RF. The ProWRAS algorithm overcomes this by using a more rigorous modeling of the convex space of the minority class \cite{LoRAS,ProWRAS}.\par 
In brief, the minority set is divided into partitions depending on the proximity of the minority samples to the majority class.
To each partition, a weight, according to its distance to the majority set is assigned. The weights are used to determinate the number of points used from each partition for synthetic sample generation.
The closer the partition to the majority set, the more synthetic samples are generated from that partition. Moreover, the weights are also used to decide the number of shadowsamples used for the convex combination to generate a synthetic sample. For a minority partition closer to the majority class, more shadowsamples are used for the convex combination, because it reduces the variance of the synthetic samples, preventing them from interfering with the majority class\cite{ProWRAS}.\par
The model offers four oversampling schemes depending on the choice of two parameters \emph{max\_{}conv} and \emph{neb\_{}conv}. \emph{max\_{}conv} selects the number of shadowsamples used to create one synthetic data point. The oversampling schemes are designed such that, with the proper choice of oversampling scheme, the algorithm can be compatible with multiple classifiers \cite{ProWRAS}.

\subsection{GAN}\label{sec:Algo:GAN}
Formally, a Generative model $G$ takes a random input $z$ and outputs a $x_G=G(z)$. Ideally, for a trained GAN, $x_G$ should follow the distribution of the original samples \cite{GAN}. The Discriminator network receives as input $x$ that can be an original sample or a synthetic sample and returns the probability of $D(x)$ of $x$ to be an original sample \cite{GAN}. These two models $G$ and $D$ compete in a two-player min-max game with value function:
\begin{equation}
    \min_G\max_DV(D,G)=\mathbb{E}_D+\mathbb{E}_G
\end{equation}
where, 
\begin{equation}
    \mathbb{E}_D=\mathbb{E}_{x \sim P_{data}(X)}[\log(D(x))]
\end{equation}
and,
\begin{equation}
    \mathbb{E}_G=\mathbb{E}_{z \sim P_{z}(L)}[\log(1-D(G(z)))]
\end{equation}

$x\in X$ are data points sampled from the data distribution $P_{data}$ and $z$ are random noises sampled from some distribution $P_z$; $\mathbb{E}_P[\cdot]$ represents expectation of a random variable over a distribution $P$ \cite{GAN}. The training procedure consists of alternating between $k$ optimizing steps for $D$ and one optimizing step for $G$ by applying a proper optimizer. Therefore, during training, $D$ is optimized to correctly classify training data and samples generated from $G$, assigning $1$ and $0$ respectively. On the other hand the generator is optimized to confuse the discriminator \cite{GAN}. Thus, the generator learns through backpropagation from each decision of the discriminator, which leads the generator to produce more and more realistic samples until the discriminator can not distinguish between real and synthetic data points anymore. In our implementation as a pre-processing step, we scale the data by a factor of $\alpha^{-1}$, where 
\[ \alpha =\max\left( 1.0, 1.1 \cdot \max_{x \in data} \|x\|_\infty \right) \]
with $\|x\|_\infty = \max \left\{ |x_i| : i \in \{ 1, 2, 3, \ldots, n\} \right\}$ is the $L^{\infty}$ norm.
The final layer uses the \emph{softsign} function to project to the output in range $\mathcal{B} = \{ (x_1, \ldots , x_n) \in \mathbb{R}^n : -1 \leq x_i \leq 1  \}$ where $n$ is the number of features in a dataset. The generated synthetic points are scaled back by multiplying with $\alpha$, after training. This step was essential due to the range of the \emph{softsign} function, $[-1,1]$, as shown in Table \ref{tbl:activationFunctions}. Both the generator and the discriminator are Multi-Layered Perceptrons (MLP). The intermediate layers of our GAN model use \emph{ReLU} as activation function. The used noise size is $\nu = 16 n$. The layer for the generator have the size $2 \nu, 4 n $ and $2 n$. The discriminator uses the layer sizes $40 n$, $20 n$ and $10 n$ with \emph{ReLU} as output function. The last layer uses the \emph{sigmoid} activation function.

\begin{table}[ht]
\scriptsize
\caption{Different activation functions used for the GAN model.}
\label{tbl:activationFunctions}
\centering
\tabularnewline
\resizebox{\columnwidth}{!}{\begin{tabular}{l |@{\hskip3pt}c@{\hskip3pt}|@{\hskip3pt} c @{\hskip3pt}}
\hline
\textbf{Name} & \textbf{Function} & \textbf{Range} \tabularnewline
\hline
  ReLU     & $f(x) = \max \{ 0, x \}$ & $0 \leq f(x)$ \tabularnewline
  softsign & $f(x) = {x / (|x| + 1)}$ & $-1 < f(x) < 1$ \tabularnewline
  sigmoid  & $f(x) = {1 / (1 + \exp (-x}))$ & $0 < f(x) < 1$ \tabularnewline
\hline
\end{tabular}}
\end{table}

\subsection{CTGAN}\label{sec:Algo:CtGAN}
CTGAN (GIT link: \url{https://github.com/sdv-dev/CTGAN}) is a synthetic data generation library for tabular datasets based on Conditional Generative Adversarial Networks. To model Non-Gaussian and multimodal continuous distributions Xu \textit{et al.} \cite{CtGAN} use three steps. 
First, they use a variational Gaussian mixture model (VGM) to estimate the number of modes for a given continuous column. Then, for each value in the selected column the probability of belonging to a specific mode is computed. In the last step, a value is sampled from that probability density and normalized for the sampled mode ({mode-specific normalization}). Each continuous value is therefore represented with a one hot encoded mode and a continuous mode-specific normalized scalar.     
Discrete data is represented using one-hot encoding \cite{CtGAN}.\par
To address challenges with imbalanced discrete data the algorithm uses a {conditional generator} and {training by sampling}. Therefore, a conditional (one-hot-encoded) vector was introduced. This vector represents the selection of one specific value for just one discrete column. Thus, the size of the vector is the sum of the cardinality of each discrete column. 
During training a column with discrete values is uniformly selected. Then, a specific value is sampled according to the PMF across all possible values (the probability mass of each value is the logarithm of its frequency in that column) \cite{CtGAN}. For a selected value in the column, the conditional vector is set accordingly. The generator is fed with this conditional vector and a noise vector consisting of random values. The loss function of the generator has been altered to additionally penalize the generator in cases where it does not produce the selected column with the selected value. To prevent mode collapse the idea of `packing' is introduced where the discriminator works on multiple samples simultaneously \cite{packing}. \par       
To train CTGAN, a list of discrete columns needs to be provided. Other columns are treated as continuous data. For our experiments, we automatized the decision, so that columns with integer data only are considered discrete, and columns with at least one fractional value are considered continuous.

\subsection{CTAB-GAN}\label{sec:Algo:CtabGAN}
Zhao \textit{et al.} \cite{CtabGAN} introduced CTAB-GAN (GIT link: \url{https://github.com/Team-TUD/CTAB-GAN}) for synthetic tabular data generation, which is also based on Conditional Generative Adversarial Networks. In addition to a Generator and Discriminator, CTAB-GAN uses an auxiliary Classifier. This Classifier (multilayer perceptron) is trained to improve semantic integrity of the synthetic data. Both Generator and Discriminator are implemented using CNNs.
The feature vector is represented by a square matrix of the size $\lceil \sqrt{n} \rceil$ where $n$ is the number of features in the vector. Missing values are padded by $0$.
To improve the Generator, two additional losses has been added: information loss (to keep the shape of the data) and classification loss (to prevent the creation of invalid data, for example, female with prostate cancer). Zhao \textit{et al.} use a \emph{conditional generator} and \emph{training-by-sampling}, which is inspired by the paper of Xu \textit{et al.} \cite{CtGAN}. The development of CTAB-GAN was motivated by the limitation of prior state-of-the-art:


\begin{itemize}
    \item \emph{Mixed data type variables}: Tabular data can contain values with special meaning or limitations.
        For example a continuous variable could have $0$-values representing \emph{not available}.
    
    \item \emph{Skewed multi-mode continuous variables}: The histogram of the data can have various peaks in different heights.
    
    \item \emph{Long tail distribution}: Most of the values are near one point and rare values are far away from that point.
\end{itemize}

A Mixed-type Encoder is proposed to deal with mixed data. A column is considered mixed if it contains both categorical and continuous values or continuous values with missing values \cite{CtabGAN}. Mixed variables are treated as value pairs, consisting of a categorical part (used also for missing values) and a continuous part. The individual parts are handled similarly in CTGAN \cite{CtabGAN}. However, these two parts build just one vector for the variable.

The concept of training by sampling introduced in CTGAN paper has been extended here to include continuous variables. During training each column has the same probability of being selected. For continuous variables a specific mode is then sampled from the probability distribution (logarithm of the frequency) of the modes \cite{CtabGAN}.  

To handle long tail distributions Zhao \textit{et al.} pre-process variables with such distribution with a logarithm transformation \cite{CtabGAN}.



\section{The ConvGeN algorithm} \label{sec:ConvGeN}
\textbf{An overview of the philosophy behind ConvGeN:} \mbox{ConvGeN} generates synthetic samples in the convex space of the minority class, much like SMOTE and its extensions or like ProWRAS and LoRAS\cite{SMOTE, LoRAS, ProWRAS}. However, a key contrasting feature of ConvGeN against such conventional linear interpolation models is that the coefficients of convex combinations are not chosen randomly, but learned by the model itself during the training, using two co-operating neural networks. LoRAS and ProWRAS, although they developed the idea of a more rigorous modeling of the convex space to improve imbalanced classification, this modeling of the convex space has to be done in a dataset-specific fashion by rigorous parameter tuning \cite{ProWRAS}. This is the reason why ProWRAS works best when the oversampling algorithm is customized for each dataset and classifier \cite{ProWRAS}. Deep co-operative learning enables ConvGeN to model the convex space automatically over learning iterations.\par
ConvGeN has a generator-discriminator architecture similar to GANs. However, there are key differences:
\begin{itemize}
    \item To begin with, the generator and the discriminator of ConvGeN, unlike conventional GANs, do not compete against each other, but cooperate among themselves. The optimization of the loss functions for the generator and the discriminator are not in contradiction to each other. ConvGeN is thus not an adversarial network.
    \item The generator for ConvGeN takes shuffled batches of minority data points as input and generates convex linear combinations of input data points as synthetic samples. For conventional GANs, the generator takes random noise as input and generates realistic synthetic samples that are not necessarily convex combinations of the original samples. 
    \item  The discriminator of ConvGeN learns to classify the synthetic points created by the generator against a randomly selected and shuffled batch from the majority class. The generator learns the coefficients of convex combinations for synthetic data generation for every minority class data neighborhood in such a way, that it also leads to an improved classification for the discriminator. Note that, for this model, the discriminator can itself be used as a minority-majority classifier. For conventional GANs this is not the purpose of the discriminator. There, the discriminator differentiates whether the given sample is real or synthetic.
\end{itemize}

To ensure that ConvGeN can learn the minority data manifold in every small minority data neighborhood, it is trained using neighborhood data batches, that is, each batch consists of minority data from a single minority neighborhood. This is necessary to avoid the limitation of traditional GANs that learn some portions of the data space and tend to generate synthetic samples relevant only to the learned portions \cite{modecollapse}. ConvGeN will be able to learn the relevant convex space for all minority data neighborhoods and therefore generate synthetic data at will from the convex space of any minority neighborhood of choice.\par
\textbf{Mathematical formalization of the ConvGeN model:}
For a given data point $x \in X$, where $X$ is the minority class, we can denote the $k$-neighborhood of $x \in X$ as $N(x)\subseteq X$. We assume that $N(x)$ (in short, we will sometimes call this $N$) has been randomly shuffled. The Generator $G$ will take as input $N(x)$ and map it to the convex space of $N(x)$, which we can call $\mathbb{C}_{N(x)}$. Random shuffling of $N(x)$ is crucial, as it can increase the number of training instances, if there is a less number of samples. For example, with $k=10$, one can generate $10!=1024$ combinations that can be treated as individual training samples. Also, ideally, when we talk about learning the coefficients of convex combinations, customized to each minority neighborhood, it means that these coefficients are learned for any shuffle of the neighborhood since we assume no sense of order among data points in a minority data neighborhood.\par
 \[G:N(x)\subseteq X \rightarrow C_N(x)\subseteq \mathbb{C}_{N(x)},\] where $C_N(x)$ (in short, we will call this $C_N$ sometimes) is a convex combination of the samples in $N(x)$. Note that here we impose a constraint on $G$ generating synthetic samples from the convex space of the original data neighborhood. This ensures that the distribution of generated synthetic samples, if considered as random variables from the corresponding minority neighborhood, have identical mean as the distribution of the original samples. This is evident as given identically distributed random variables $X_1,\dots,X_n$, with mean $\mathbb{E}[X_i]=\mu$ their convex combination $Y=\sum_{i=1}^n\alpha_iX_i$ has the mean \cite{LoRAS}
\[\mathbb{E}[Y]=\mathbb{E}\bigg[\sum_{i=1}^n\alpha_iX_i\bigg]=\mu\sum_{i=1}^n\alpha_i=\mu=\mathbb{E}[X_i].\] Therefore, the objective of the ConvGeN model is to learn the optimal coefficients for each data neighborhood, so that the synthetic samples are distinct enough from the majority class, while also being similar to the original samples. In simple words, the discriminator is designed to ensure that the variance of the synthetic samples optimally learned, in the sense that the synthetic samples do not spread too much outside the neighborhood. The discriminator $D$, receives as input the synthetic samples generated from a particular minority neighborhood $N(x)$ by $G$ and learns to classify them against a random shuffled batch of majority samples $M$, so that $|M|=|C_N|$. \par
\[ D:C_N\oplus M\subset X^c \rightarrow[0,1],\] 
where $\oplus$ denotes a concatenation and $X^c$, is the majority class, which is the complement of $X$. $D$ thus ensures that the synthetic samples generated as convex combinations of samples in $N(x)$ are distinct from the majority data space. $D$ can pass on its knowledge to $G$ during the interactive training process, such that $G$ learns to choose appropriate convex coefficients for synthetic sampling generation within $N(x)$. Learning the proper convex coefficients by $G$ can be intuitively interpreted as if the model makes sure, the distribution of generated samples, in addition to having an identical mean as the distribution of the original samples, also has an optimal variance that assists the classification task. \par

The discriminator thus receives a sample, which can either be a majority sample or a synthetic sample from the convex minority space. We used binary cross-entropy loss to train the discriminator, which is defined as:
\[L_D=-\sum_{c\in C_N, y\in M} \log(D(c))+\log(1-D(y))\]
There is no separate training for the generator, as we assume that the convex combinations of minority data points generated by $G$ are already similar to the original samples (distributions share identical means). The weights of $G$ are updated when the whole ConvGeN model is trained.
The ConvGeN model uses the mean-squared error as a loss function:
\[L_{\text{ConvGeN}}=\text{MSE}(Y_{N\oplus M},\hat{Y}_{N\oplus M})\]
where $Y_{N\oplus M}$ are the labels of the concatenated minority neighborhood batch (shuffled) $N$ and the random majority batch $M$ (shuffled), while $\hat{Y}_{N\oplus M}$ are the corresponding predicted labels.  We provide the pseudocode for ConvGeN in Algorithms \ref{algo:convGan:train} and \ref{algo:convGan:train:discriminator}.
\par 

\textbf{Detailed architecture, parameters, and working mechanism of ConvGeN:} An adversarial learning process can be viewed as a min-max game, where the discriminator seeks to maximize the quantity $-L(D)$ whereas the generator seeks to achieve the reverse. For a conventional GAN, the discriminator tries to minimize its own loss ($L_D$), whereas the generator tries otherwise. The competition between the generator and the discriminator leads to synthetic samples with distribution similar to the original data. However, in the case of ConvGeN, $D$ and $G$ do not compete, but cooperate. $G$ does not try to minimize $-L_D$ in ConvGeN, as it happens in case of GANs. $D$ receives the original labels of the samples. The aim of ConvGeN is to learn the subset $S$ of the convex space of the minority class $\mathbb{C}_{X}$, such that sampling from that subset can help the discriminator to minimize $L_D$, meaning that the classification task is facilitated by the synthetic data generation process. The underlying assumption is of course, that the generated synthetic samples from $S$ guarantees a reasonable classification model. Note that, given the performance of ProWRAS and LoRAS, this is a reasonable assumption to make \cite{LoRAS, ProWRAS}. Figure \ref{fig:ConvGeN} depicts the architecture of ConvGeN.\par

We kept the architecture of ConvGeN relatively simple refraining from making the neural networks too deep. In the architecture of $G$ in ConvGeN, the input layer has $f$ nodes, $f$ being the number of features. At a time, it receives a shuffled batch of \var{neb} original minority samples. The specialty is that each input batch is a neighborhood $N$ (\var{neb} number of the closest data points) of some minority class data point. Thus, the dimension of the input is \var{neb}$\times f$. After the input layer a 1-$D$ convolution layer is used to reduce the dimension of the input to $k'\times f$, where $k'< \var{neb}$. Then the output of the convolution layer is flattened and reshaped to obtain a matrix $K_N$ of dimension \var{neb}$\times$\var{gen}, such that each column of $K_N$ sums to 1 and the entries of $K_N$ are non negative (this is achieved using ReLU activation). Then a residual connection with the input $N$ yields the matrix $C_N=K_N^T\times N$ ($K_N^T$ being the transpose of $K_N$), whose every row is a convex combination of the minority neighborhood data points in $N$.\par
The architecture of $D$ in ConvGeN is straightforward. The input layer has $f$ nodes, $f$ being the number of features. At each training step, it receives a batch of \var{gen}=\var{neb} synthetic samples $C_N$ from $G$ concatenated with a batch of \var{gen}=\var{neb} majority samples $M$ as input. We used three dense layers of size $250$, $125$, and $75$ respectively, while the output layer consists of two nodes for two classes. The true labels of the inputs are also revealed to $D$, such that, synthetic samples  from $G$ are labeled as minority data. The architecture of ConvGeN is shown in Figure \ref{fig:ConvGeN}.\par
\textbf{Computational implementation:}
In terms of computational implementation, the algorithm has four parameters: \var{neb}: size of the shuffled minority data neighborhood that will be used for modeling the convex space of the minority class, \var{gen}: number of generated synthetic points in form of convex combinations of the minority neighborhood batch by generator at a time, \var{disc\_train\_count}: number of extra training steps for the discriminator before updating the generator once, and \var{maj\_proximal}: an indicator for sampling the majority batches either from the entire majority class or a borderline subset of the majority class, that is, a subset of the majority class, which is relatively closer to the minority class.\par
We have considered \var{gen}$=$\var{neb} for our model. The parameter \var{maj\_proximal}, if set to be true, the algorithm first detects a subset of the majority class that is closest to the minority class, which we call the proximal majority set. This is done by calculating the \var{neb} nearest neighbors of each minority sample in the majority class (for the training data) and then taking the union of those majority class samples. The proximal majority set is the union of the majority class nearest neighbors of all minority samples/data points. We tested four different settings of the ConvGeN model, shown in Table \ref{tbl:ConvGeN_models}. The motivation behind this, is the practical usability of the model to handle larger tabular datasets, which we have discussed later in more detail (see Section \ref{sec: results}: \textbf{ConvGeN(min,maj) model produces the best classification scores among different ConvGeN parameter settings}). However, for our comparative benchmarking study on small tabular imbalanced datasets, we have consistently used the ConvGeN (min,maj) setting. \par
Note that, in conventional deep learning, we use the term \textit{epoch} to denote one training step where a deep leaning model is trained on the entire training data once. However, for ConvGeN, we use minority neighborhoods as input for the generator. Therefore, in case of ConvGeN, we use the term \textit{neighborhood epoch} for a training step. During a training step $G$ receives all minority neighborhoods as one input batch (each time concatenated with a randomly shuffled majority batch of equal size), that is for $x\in X$, $N(x)$ is passed through $G$ once. That is why in Algorithm \ref{algo:convGan:train}, we see that the outer for loop runs in the range $(1,$\var{neb\_epochs}$)$. The entire training procedure is coded in pseudocode for Algorithm \ref{algo:benchmark}. We used {\fontfamily{pcr}\selectfont  Python (V 3.8.10)} for implementations of the algorithms and experiments.

\begin{table}[ht]
\scriptsize
\caption{Different ConvGeN models that we have tested in our study, according to parameter settings. These parameter settings for ConvGeN can be retrieved as standard parameter settings and can be used as a starting point to optimize the model for future works.}
\label{tbl:ConvGeN_models}
\centering
\tabularnewline
\resizebox{\columnwidth}{!}{\begin{tabular}{l |@{\hskip3pt}c@{\hskip3pt}}
\hline
\textbf{model name} & \textbf{parameter settings} \tabularnewline
\hline
  ConvGeN(5,maj)    & \var{neb}$=5$, \var{maj\_proximal}$=$\var{False}
  \tabularnewline
  ConvGeN(min,maj) & \var{neb}$=|X|$, \var{maj\_proximal}$=$\var{False}
  \tabularnewline
  ConvGeN(5,prox)  & \var{neb}$=5$, \var{maj\_proximal}$=$\var{True}
  \tabularnewline
  ConvGeN(min,prox)  & \var{neb}$=|X|$, \var{maj\_proximal}$=$\var{True}
  \tabularnewline
\hline
\end{tabular}}
\end{table}

\begin{algorithm*}[!htbp]
\small
    \SetArgSty{textnormal}
    \caption{ConvGeN train algorithm}
    \label{algo:convGan:train}
    \SetKwInOut{In}{Inputs}
    \In{\newline
        \vspace{-0.5em}
        \begin{flushleft}
        \begin{tabular}{ l l }
            \var{data} & Labeled data with minority and majority class.
        \end{tabular}
        \end{flushleft}
    }
    \SetKwInOut{Parameter}{Parameters}
    \Parameter{
        \begin{flushleft}
        \vspace{-0.5em}
        \begin{tabular}{ l l l }
            \var{disc\_train\_count} & $\geq 0$ & Extra training steps for the discriminator before updating generator once. (used value: $5$) \\
            \var{neb} & $\geq 2$ & Size of the minority neighborhood used to model convex minority data space. \\
            \var{gen} & $\geq \var{neb}$ & Number of generated synthetic points by generator at a time. (used value: $\var{gen} = \var{neb}$) \\
        \end{tabular}
        \end{flushleft}
    }

    \SetKwFor{For}{For}{do}{endfor}
    \SetKwIF{If}{ElseIf}{Else}{If}{then}{else If}{else}{endif}
    
    \vspace{1em}

    \textbf{Function} \var{train}(\var{data})
    
    \Begin{
        $\var{labels} \leftarrow \big( \underbrace{(1, 0), ... ,(1, 0)}_{gen} , \underbrace{(0, 1), ... ,(0, 1)}_{gen} \big)$
        \vspace{3pt}

        \For{$1 \ldots \var{neb\_epochs}$}{
            \vspace{5pt}
            \For{$1 \ldots$ \var{disc\_train\_count}}{
                \For{$x \in$ minority class}{
                    $\var{train\_discriminator}(\var{x}, \var{labels}, \var{data})$
                }
            }
            \vspace{1em}
            
            \For{$x \in$ minority class}{
                (\var{concat\_sample}, \var{min\_batch}, \var{maj\_batch})
                $\leftarrow \var{train\_discriminator}(\var{x}, \var{labels}, \var{data})$
                \vspace{.5em}
                
                train generator by using
                $\var{discriminator}\big( (\var{generator}(\var{min\_batch}) ~\oplus~ \var{maj\_batch}) \big)$
                \\ with \var{concat\_sample} $\mapsto$ \var{labels}
            }
        }
    }
\end{algorithm*}

\begin{algorithm*}[!htbp]
\small
    \SetArgSty{textnormal}
    \caption{ConvGeN train algorithm (discriminator)}
    \label{algo:convGan:train:discriminator}
    \SetKwInOut{In}{Inputs}
    \In{\newline
        \vspace{-0.5em}
        \begin{flushleft}
        \begin{tabular}{ l l }
            \var{x} & Minority class data point.\\
            \var{labels} & List of $2 \var{gen}$ labels for the training step. 
                (1,0): item is in minority class. (0,1): item is in majority class.\\
            \var{data} & Labeled data with minority and majority class.
        \end{tabular}
        \end{flushleft}
    }
    \SetKwInOut{Parameter}{Parameters}
    \Parameter{
        \begin{flushleft}
        \vspace{-0.5em}
        \begin{tabular}{ l l l }
            \var{neb} & $\geq 2$ & Size of the neighborhood used to create one synthetic point. \\
            \var{gen} & $\geq \var{neb}$ & Number of generated synthetic points per step. (used value: $gen = neb$) \\
            \var{maj\_proximal} & ~ & \var{True}: Train with \var{gen} points from nearest neighborhood for \var{x} in majority class. \\
            ~ & ~ & \var{False} Train with \var{gen} random points in the majority class.
        \end{tabular}
        \end{flushleft}
    }

    \SetKwFor{For}{For}{do}{endfor}
    \SetKwIF{If}{ElseIf}{Else}{If}{then}{else If}{else}{endif}
    
    \vspace{1em}

    \textbf{Function} \var{train\_discriminator}(\var{x}, \var{labels}, \var{data})
    
    \Begin{
        \var{min\_batch} $\leftarrow$ Set of \var{neb} nearest points to \var{x} in the minority class in randomized order.
        
        \var{conv\_samples} $\leftarrow$ generate \var{gen} points using the neighborhood of \var{x}.
        \vspace{.5em}
        
        \uIf{maj\_proximal}{
            \var{maj\_batch} $\leftarrow$ list of \var{gen} points in the nearest neighborhood of \var{x} in the majority class in random order.
        }\Else{
            \var{maj\_batch} $\leftarrow$ list of \var{gen} random points in the majority class.
        }
        \vspace{.5em}
        
        \var{concat\_sample} $\leftarrow (\var{conv\_samples} \oplus \var{maj\_batch})$
        \vspace{.5em}
        
        train discriminator with \var{concat\_sample} $\mapsto$ \var{labels}
        \vspace{.5em}
        
        return (\var{concat\_sample}, \var{min\_batch}, \var{maj\_batch})
    }
\end{algorithm*}

\section{Experimental protocols}\label{sec:case studies}

\subsection{General protocols for benchmarking}\label{Protocols for benchmarking}

As far as data science related experiments are concerned, we have closely followed the published work of Bej \textit{et al.}\ \cite{ProWRAS}. In brief, for each dataset we have used $5\times5$ stratified cross-validation. This means we have used five random shuffles of data and performed a $5$-fold cross validation on all shuffles. The average values of the evaluation metrics or performance measures, F1-score, and $\kappa$-score over all the shuffles and folds are reported. Moreover, we ensured that the test data for any given fold in each shuffle is not used for oversampling meaning that, we only used the training data for this purpose.\par 
Following the work of Bej \textit{et al.}\ \cite{ProWRAS}, we evaluated the classification model performances using F1-score and Cohen's $\kappa$-score. F1-score is the harmonic mean of precision and recall and is a well-suited measure for the overall classification performance for the minority class. Given a classification problem, the $\kappa$ measure is formally defined as:
\begin{equation*}
    \kappa=\frac{P_o-P_e}{1-P_e},
\end{equation*}
where $P_o$ is the measure of observed agreement among the chosen classifier and the ground truths of the classification problem. $P_e$ is the measure of agreement by chance, among a chosen classifier and the ground truths of the classification problem. The $\kappa$-score gives us a quantification of how good the classification is, considering both the majority and the minority class. \par

\begin{algorithm*}[!htbp]
\small
    \SetArgSty{textnormal}
    \caption{benchmark algorithm}
    \label{algo:benchmark}
    \SetKwInOut{In}{Inputs}
    \In{\newline
        \vspace{-0.5em}
        \begin{flushleft}
        \begin{tabular}{ l l }
            \var{data} & List of data points with minority and majority class.
        \end{tabular}
        \end{flushleft}
    }
    \SetKwInOut{Parameter}{Parameters}
    \Parameter{
        \begin{flushleft}
        \vspace{-0.5em}
        \begin{tabular}{ l l l }
            \var{n\_folds} & $\geq 1$ & Number of parts the data is divided to. (used value: $5$)\\
            \var{n\_shuffles} & $\geq 1$ & Number of benchmark steps to do. (used value: $5$)\\
            \var{classifiers} & ~ & Set of classifiers to evaluate the algorithm with. \\
        \end{tabular}
        \end{flushleft}
    }

    \SetKwFor{For}{For}{do}{endfor}
    \SetKwIF{If}{ElseIf}{Else}{If}{then}{else If}{else}{endif}
    
    \vspace{1em}

    \textbf{Function} \var{benchmark}(\var{gan}, \var{data})
    
    \Begin{
        Shuffle \var{data}.    
        \vspace{.5em}

        \For{$1 \ldots \var{n\_shuffles}$}{
            Shuffle \var{data}.    
            \vspace{.5em}
    
            $\var{fold}_1, ... , \var{fold}_{n\_folds} \leftarrow $ split \var{data} equally to \var{n\_folds} parts. The proportion of each class in each part has to be the same proportion of each class in the entire dataset. The last part might contain less data-points.
            \vspace{.5em}
            
            \For{$k = 1 \ldots n\_folds$}{
                $\var{data\_test} \leftarrow \var{fold}_k$
                
                $\var{data\_train} \leftarrow \var{fold}_1 \cup \ldots \cup \var{fold}_{k-1} \cup \var{fold}_{k+1} \cup \ldots \cup \var{fold}_{\var{n\_folds}}$
                \vspace{.5em}

                reset ConvGeN
                \vspace{.5em}

                train ConvGeN with \var{data\_train}
                \vspace{.5em}

                $\var{n\_synthetic} \leftarrow$ (size of majority class in \var{data\_train}) $-$ (size of minority class in \var{data\_train})

                $\var{data\_synthetic} \leftarrow$ generate \var{n\_synthetic} points with ConvGeN

                $\var{data\_train\_full} \leftarrow \var{data\_train} \cup \var{data\_synthetic}$
                \vspace{.5em}

                \For{$\var{c} \in \var{classifiers}$}{
                    train \var{c} with \var{data\_train\_full}

                    predict \var{data\_test} with \var{c}

                    score the prediction with $f_1$-score and $\kappa$-score
                }
            }
        }
    }
\end{algorithm*}

\subsection{Datasets and Algorithms used}\label{Benchmarking datasets}

\begin{table}[ht]
\scriptsize
\caption{Table showing statistics of datasets used for the benchmarking study.}
\centering
\tabularnewline
\resizebox{\columnwidth}{!}{\begin{tabular}{l |@{\hskip3pt}c@{\hskip3pt}|@{\hskip3pt} c @{\hskip3pt}|@{\hskip3pt} c@{\hskip3pt}|@{\hskip3pt} c@{\hskip3pt}}
\hline\label{tbl:methods:datasets}

\textbf{Dataset} & \textbf{Features} & \textbf{Total data size} & \textbf{Minority class size} & \textbf{Imbalance Ratio} \tabularnewline
\hline
 abalone 17 vs 7 8 9 10          &   8 &   2338 &   58 & 39.31 \tabularnewline
 abalone (9-18)                  &   8 &    731 &   42 & 16.40 \tabularnewline
 car good                        &   6 &   1728 &   69 & 24.04 \tabularnewline
 car vgood                       &   6 &   1728 &   65 & 25.58 \tabularnewline
 flare-F                         &  11 &   1066 &   43 & 23.79 \tabularnewline
 hypothyroid                     &  25 &   3163 &  151 & 19.95 \tabularnewline
 kddcup guess passwd vs satan    &  38 &   1642 &   53 & 29.98 \tabularnewline
 kr-vs-k 3 vs 11                 &   6 &   2935 &   81 & 35.23 \tabularnewline
 kr-vs-k zero-one vs draw        &   6 &   2901 &  105 & 26.63 \tabularnewline
 shuttle 2 vs 5                  &   9 &   3316 &   49 & 66.67 \tabularnewline
 winequality red 4               &  11 &   1599 &   53 & 29.17
 \tabularnewline
 yeast4                          &  10 &   1484 &   51 & 28.10
 \tabularnewline
 yeast5                          &  10 &   1484 &   44 & 32.73 \tabularnewline
 yeast6                          &  10 &   1484 &   35 & 41.40
 \tabularnewline
\hline

\end{tabular}}
\end{table}

Our benchmarking datasets are a subset of the $104$ publicly available imbalanced datasets used for the benchmarking studies in Kovács \textit{et al.} \cite{Comparison}. In Table \ref{tbl:methods:datasets}, we show the statistics for the relevant datasets.

Out of $104$ publicly available imbalanced datasets, satisfying all the following three criteria, to ensure that our choice of datasets for the studies are impartial, following the work of Bej \textit{et al.}\ \cite{ProWRAS}. The criteria are:  
\begin{itemize}
    \item Datasets with an imbalance ratio of at least $15:1$ were chosen. This is to ensure that the performance of the compared oversampling algorithms are tested, particularly on datasets with high imbalance following Bej \textit{et al.}\ \cite{ProWRAS}.\\
    \item We choose datasets with a minority class samples in the range $35$ to $151$. For datasets with very less minority class samples, classifier performances are often affected by high stochastically and the results are often statistically unreliable \cite{ProWRAS}. While too many samples provide the generative networks too many training samples for synthetic sample generation, which is a case outside the scope of this study, we are investigating only tabular datasets of small size.\\
    \item Datasets with at most $3500$ samples were used as we are investigating small tabular data.
\end{itemize}

Recall that, we chose six oversampling algorithms for our benchmarking study, which are discussed in more detail in \ref{sec:Algo}. Below we discuss the rationale behind the choice of the oversampling algorithms for our benchmarking study.
\begin{itemize}
    \item The repeater model was chosen as the baseline for our benchmarking experiments. In common articles on oversampling datasets without oversampling are conventionally chosen as the baseline, which obviously perform worse than the proposed models. The rationale behind our choice of this baseline is that, even without the availability of any benchmarking algorithm, a modeler can balance the classes by simply repeatedly sampling existing minority samples with replacement, the effect of which has been unexplored.
    \item The traditional GAN model is the pioneer of deep generative modeling, for which it is important to compare ConvGeN against it. 
    \item Moreover, we used two variants CTGAN (2019) and CTAB-GAN (2021) of the GAN model that are not only recent, but also extensively cited.
    \item We used the ProWRAS algorithm for benchmarking, since it is a state-of-the-art model for convex space learning, a philosophy transferred from ProWRAS into ConvGeN.
\end{itemize}

ProWRAS has been used with default settings. The CTGAN model was trained for $300$ (default value) epochs and columns with all integer values were marked as discrete columns while for the rest of the model parameters the default settings were used. The CTAB-GAN model was also trained for $300$ (default value) epochs, while for the rest of the model parameters the default settings were used. GAN, CTGAN, and CTAB-GAN were trained only on the minority class. The GAN model that we used was also trained for $300$ epochs maintaining consistency with CTGAN and CTAB-GAN. Finally, the ConvGeN model was trained for $10$ neighborhood epochs for four different parameter settings as shown in Table \ref{tbl:ConvGeN_models}. However, for comparative studies we only show the ConvGeN(min,maj) model.\par
We used four commonly used traditional machine learning classifiers, following the work of Bej \textit{et al.}\ \cite{ProWRAS}. The \textit{Logistic Regression} (LR) classifier was chosen because it is a relatively fast model. The \textit{k-Nearest Neighbors} (kNN) is also observed to perform well for imbalanced datasets in the benchmarking studies of Bej \textit{et al.}\ \cite{LoRAS, ProWRAS}. Moreover, we used \textit{Gradient Boosting} (GB) and \textit{Random Forest} (RF) as ensemble models, as they are seen to perform well with the ProWRAS algorithm \cite{ProWRAS}. Although ensemble models can perform well for classification on small tabular imbalanced data, especially with a large number of features, there is evidence that ensemble models are not robust \cite{overfitting_GB,overfitting_RF}. For instance, for GB, this occurs due to the working principle of Boosting models where erroneous decisions on data points, which are difficult to classify, are corrected through multiple iterations by making the model learn these data points multiple times, hence over emphasizing on these data points. One could therefore argue that, simpler models like kNN, albeit under-performing, can be safer to use for decision-making on unseen data. Considering this, we investigated multiple classifiers in our studies (LR, kNN, RF and GB). We used default parameters as recommended in {\fontfamily{pcr}\selectfont scikit-learn (V 0.21.2)} documentation.\par
Moreover, recall that the discriminator of a trained ConvGeN model can itself act as a classifier on the data generated by ConvGeN. This is why, for ConvGeN-generated synthetic samples, we use the discriminator of the trained ConvGeN model (DoC), also as a classifier. After the training of ConvGeN is finished, we can use the generator to generate synthetic samples from each neighborhood of the convex space. These synthetic samples are used to produce a balanced dataset, which is used to retrain the discriminator for some epochs (we used $10$ epochs). The purpose of this is that, the discriminator, while training of ConvGeN, only sees random batches of the majority class of the training data. This retraining ensures that the retrained discriminator includes all majority class samples in its training.

\begin{table*}[htbp]\scriptsize\caption{Table showing F1-score/$\kappa$- score for compared oversampling strategies for LR classifier on 14 benchmarked imbalanced datasets. Models with the best performances are marked in bold for each dataset. ConvGeN yields the best average $\kappa$- score and F1-Score. }\label{tab:results:LR:A}\centering\tabularnewline\begin{tabular}{l|@{\hskip3pt}c@{\hskip3pt}|@{\hskip3pt}c@{\hskip3pt}|@{\hskip3pt}c@{\hskip3pt}|@{\hskip3pt}c@{\hskip3pt}|@{\hskip3pt}c@{\hskip3pt}|@{\hskip3pt}c@{\hskip3pt}}\hline
\textbf{dataset ($f_1~$score$~/~\kappa~$score)} & \textbf{Repeater} & \textbf{ProWRAS} & \textbf{GAN} & \textbf{CTGAN} & \textbf{CTAB GAN} & \textbf{ConvGeN(min,maj)}
\tabularnewline
\hline
abalone 17 vs 7 8 9 10 &  0.275  /  0.244  &  \textbf{0.337}  /  \textbf{0.311}  &  0.304  /  0.276  &  0.221  /  0.191  &  0.200  /  0.164  &  \textbf{0.337}  /  0.310 
\tabularnewline
abalone9 18 &  0.458  /  0.409  &  0.570  /  0.538  &  0.462  /  0.418  &  0.341  /  0.288  &  0.389  /  0.332  &  \textbf{0.575}  /  \textbf{0.544} 
\tabularnewline
car good &  0.106  /  0.035  &  0.100  /  0.029  &  \textbf{0.162}  /  \textbf{0.101}  &  0.106  /  0.035  &  0.109  /  0.038  &  0.109  /  0.038 
\tabularnewline
car vgood &  0.362  /  0.320  &  0.392  /  0.354  &  0.377  /  0.337  &  0.331  /  0.287  &  0.354  /  0.311  &  \textbf{0.422}  /  \textbf{0.387} 
\tabularnewline
flare F &  0.263  /  0.210  &  0.278  /  0.228  &  0.272  /  0.223  &  0.287  /  0.236  &  0.293  /  0.244  &  \textbf{0.323}  /  \textbf{0.279} 
\tabularnewline
hypothyroid &  0.358  /  0.305  &  \textbf{0.399}  /  0.352  &  0.393  /  \textbf{0.370}  &  0.311  /  0.252  &  0.317  /  0.260  &  0.378  /  0.328 
\tabularnewline
kddcup guess passwd vs satan &  0.997  /  0.996  &  0.996  /  0.996  &  0.981  /  0.980  &  \textbf{1.000}  /  \textbf{1.000}  &  \textbf{1.000}  /  \textbf{1.000}  &  0.995  /  0.995 
\tabularnewline
kr vs k three vs eleven &  0.942  /  0.941  &  \textbf{0.965}  /  0.964  &  0.964  /  0.964  &  0.785  /  0.777  &  0.848  /  0.843  &  \textbf{0.965}  /  \textbf{0.965} 
\tabularnewline
kr vs k zero one vs draw &  0.650  /  0.632  &  0.745  /  0.733  &  0.725  /  0.712  &  0.575  /  0.552  &  0.583  /  0.561  &  \textbf{0.836}  /  \textbf{0.830} 
\tabularnewline
shuttle 2 vs 5 &  0.998  /  0.998  &  0.996  /  0.996  &  \textbf{1.000}  /  \textbf{1.000}  &  0.984  /  0.984  &  0.967  /  0.967  &  0.996  /  0.996 
\tabularnewline
winequality red 4 &  0.113  /  0.056  &  \textbf{0.143}  /  \textbf{0.090}  &  0.077  /  0.044  &  0.127  /  0.080  &  0.109  /  0.051  &  0.135  /  0.080 
\tabularnewline
yeast4 &  0.236  /  0.188  &  0.243  /  0.197  &  \textbf{0.307}  /  \textbf{0.269}  &  0.216  /  0.168  &  0.227  /  0.178  &  0.277  /  0.235 
\tabularnewline
yeast5 &  0.574  /  0.555  &  0.591  /  0.574  &  0.595  /  0.577  &  0.522  /  0.500  &  0.511  /  0.488  &  \textbf{0.620}  /  \textbf{0.604} 
\tabularnewline
yeast6 &  0.242  /  0.210  &  0.294  /  0.265  &  \textbf{0.375}  /  \textbf{0.351}  &  0.317  /  0.290  &  0.242  /  0.210  &  0.353  /  0.328 
\tabularnewline
\hline Average &  0.470  /  0.436  &  0.504  /  0.473  &  0.500  /  0.473  &  0.437  /  0.403  &  0.439  /  0.403  &  \textbf{0.523}  /  \textbf{0.494} 
\tabularnewline
\hline\end{tabular}\end{table*}

\begin{table*}[htbp]\scriptsize\caption{Table showing F1-score/$\kappa$- score for compared oversampling strategies for RF classifier on 14 benchmarked imbalanced datasets. ConvGeN yields the second-best average $\kappa$- score and F1-Score. Models with the best performances are marked in bold for each dataset. The Repeater model only outperforms the ConvGeN model strictly for 5 out of 14 datasets for both F1-score and $\kappa$- score.}\label{tab:results:RF:A}\centering\tabularnewline\begin{tabular}{l|@{\hskip3pt}c@{\hskip3pt}|@{\hskip3pt}c@{\hskip3pt}|@{\hskip3pt}c@{\hskip3pt}|@{\hskip3pt}c@{\hskip3pt}|@{\hskip3pt}c@{\hskip3pt}|@{\hskip3pt}c@{\hskip3pt}}\hline
\textbf{dataset ($f_1~$score$~/~\kappa~$score)} & \textbf{Repeater} & \textbf{ProWRAS} & \textbf{GAN} & \textbf{CTGAN} & \textbf{CTAB GAN} & \textbf{ConvGeN(min,maj)}
\tabularnewline
\hline
abalone 17 vs 7 8 9 10 &  0.190  /  0.180  &  0.169  /  0.161  &  0.256  /  0.242  &  0.239  /  0.220  &  \textbf{0.329}  /  \textbf{0.310}  &  0.280  /  0.268 
\tabularnewline
abalone9 18 &  0.322  /  0.297  &  0.367  /  0.340  &  0.319  /  0.286  &  \textbf{0.395}  /  \textbf{0.359}  &  0.376  /  0.333  &  0.348  /  0.326 
\tabularnewline
car good &  \textbf{0.899}  /  \textbf{0.895}  &  0.581  /  0.572  &  0.772  /  0.765  &  0.533  /  0.504  &  0.804  /  0.798  &  0.659  /  0.650 
\tabularnewline
car vgood &  \textbf{0.985}  /  \textbf{0.984}  &  0.914  /  0.911  &  0.949  /  0.947  &  0.812  /  0.803  &  0.959  /  0.957  &  0.929  /  0.927 
\tabularnewline
flare F &  0.222  /  0.178  &  0.095  /  0.074  &  0.127  /  0.107  &  \textbf{0.294}  /  \textbf{0.245}  &  0.121  /  0.102  &  0.120  /  0.099 
\tabularnewline
hypothyroid &  \textbf{0.800}  /  \textbf{0.791}  &  0.725  /  0.714  &  0.777  /  0.767  &  0.774  /  0.760  &  0.777  /  0.767  &  0.798  /  0.788 
\tabularnewline
kddcup guess passwd vs satan &  \textbf{1.000}  /  \textbf{1.000}  &  \textbf{1.000}  /  \textbf{1.000}  &  0.994  /  0.994  &  \textbf{1.000}  /  \textbf{1.000}  &  \textbf{1.000}  /  \textbf{1.000}  &  \textbf{1.000}  /  \textbf{1.000} 
\tabularnewline
kr vs k three vs eleven &  \textbf{1.000}  /  \textbf{1.000}  &  0.991  /  0.991  &  0.992  /  0.992  &  0.950  /  0.948  &  0.995  /  0.995  &  0.992  /  0.992 
\tabularnewline
kr vs k zero one vs draw &  0.946  /  0.944  &  0.939  /  0.936  &  \textbf{0.947}  /  \textbf{0.945}  &  0.843  /  0.836  &  0.937  /  0.935  &  0.946  /  0.944 
\tabularnewline
shuttle 2 vs 5 &  \textbf{1.000}  /  \textbf{1.000}  &  \textbf{1.000}  /  \textbf{1.000}  &  \textbf{1.000}  /  \textbf{1.000}  &  \textbf{1.000}  /  \textbf{1.000}  &  \textbf{1.000}  /  \textbf{1.000}  &  \textbf{1.000}  /  \textbf{1.000} 
\tabularnewline
winequality red 4 &  0.014  /  0.009  &  0.005  /  -0.002  &  0.024  /  0.018  &  0.153  /  0.125  &  \textbf{0.173}  /  \textbf{0.133}  &  0.086  /  0.055 
\tabularnewline
yeast4 &  0.320  /  0.306  &  \textbf{0.329}  /  \textbf{0.313}  &  0.284  /  0.272  &  0.284  /  0.252  &  0.311  /  0.300  &  0.321  /  0.304 
\tabularnewline
yeast5 &  0.720  /  0.712  &  0.706  /  0.698  &  \textbf{0.758}  /  \textbf{0.751}  &  0.647  /  0.634  &  0.740  /  0.733  &  0.738  /  0.730 
\tabularnewline
yeast6 &  0.518  /  0.508  &  0.570  /  0.562  &  0.493  /  0.485  &  0.453  /  0.437  &  0.530  /  0.522  &  \textbf{0.575}  /  \textbf{0.567} 
\tabularnewline
\hline Average &  0.638  /  0.629  &  0.599  /  0.591  &  0.621  /  0.612  &  0.598  /  0.580  &  \textbf{0.647}  /  \textbf{0.635}  &  0.628  /  0.618 
\tabularnewline
\hline\end{tabular}\end{table*}

\begin{table*}[htbp]\scriptsize\caption{Table showing F1-score/$\kappa$- score for compared oversampling strategies for GB classifier on 14 benchmarked imbalanced datasets. Models with the best performances are marked in bold for each dataset. ConvGeN yields the best average $\kappa$- score and the second-best F1-Score. The Repeater model only outperforms the ConvGeN model strictly for 4 out of 14 datasets for both F1-score and $\kappa$- score.}\label{tab:results:GB:A}\centering\tabularnewline\begin{tabular}{l|@{\hskip3pt}c@{\hskip3pt}|@{\hskip3pt}c@{\hskip3pt}|@{\hskip3pt}c@{\hskip3pt}|@{\hskip3pt}c@{\hskip3pt}|@{\hskip3pt}c@{\hskip3pt}|@{\hskip3pt}c@{\hskip3pt}}\hline
\textbf{dataset ($f_1~$score$~/~\kappa~$score)} & \textbf{Repeater} & \textbf{ProWRAS} & \textbf{GAN} & \textbf{CTGAN} & \textbf{CTAB GAN} & \textbf{ConvGeN(min,maj)}
\tabularnewline
\hline
abalone 17 vs 7 8 9 10 &  \textbf{0.333}  /  \textbf{0.312}  &  0.323  /  0.304  &  0.202  /  0.194  &  0.234  /  0.219  &  0.204  /  0.198  &  0.291  /  0.278 
\tabularnewline
abalone9 18 &  0.383  /  0.346  &  \textbf{0.399}  /  \textbf{0.362}  &  0.344  /  0.319  &  0.396  /  \textbf{0.362}  &  0.343  /  0.320  &  0.318  /  0.292 
\tabularnewline
car good &  0.839  /  0.831  &  0.849  /  0.843  &  0.849  /  0.843  &  0.533  /  0.504  &  0.840  /  0.834  &  \textbf{0.864}  /  \textbf{0.858} 
\tabularnewline
car vgood &  0.938  /  0.935  &  0.971  /  0.970  &  0.978  /  0.977  &  0.812  /  0.803  &  0.980  /  0.979  &  \textbf{0.982}  /  \textbf{0.982} 
\tabularnewline
flare F &  0.285  /  0.237  &  0.132  /  0.111  &  0.231  /  0.214  &  \textbf{0.315}  /  \textbf{0.268}  &  0.225  /  0.208  &  0.181  /  0.164 
\tabularnewline
hypothyroid &  0.781  /  0.768  &  0.795  /  0.785  &  0.808  /  0.799  &  0.774  /  0.760  &  \textbf{0.810}  /  \textbf{0.801}  &  0.800  /  0.789 
\tabularnewline
kddcup guess passwd vs satan &  \textbf{1.000}  /  \textbf{1.000}  &  \textbf{1.000}  /  \textbf{1.000}  &  0.986  /  0.986  &  \textbf{1.000}  /  \textbf{1.000}  &  \textbf{1.000}  /  \textbf{1.000}  &  \textbf{1.000}  /  \textbf{1.000} 
\tabularnewline
kr vs k three vs eleven &  \textbf{0.995}  /  \textbf{0.995}  &  \textbf{0.995}  /  \textbf{0.995}  &  \textbf{0.995}  /  \textbf{0.995}  &  0.950  /  0.949  &  \textbf{0.995}  /  \textbf{0.995}  &  \textbf{0.995}  /  \textbf{0.995} 
\tabularnewline
kr vs k zero one vs draw &  0.944  /  0.942  &  0.959  /  0.958  &  0.959  /  0.958  &  0.825  /  0.818  &  0.955  /  0.953  &  \textbf{0.972}  /  \textbf{0.971} 
\tabularnewline
shuttle 2 vs 5 &  \textbf{1.000}  /  \textbf{1.000}  &  \textbf{1.000}  /  \textbf{1.000}  &  \textbf{1.000}  /  \textbf{1.000}  &  \textbf{1.000}  /  \textbf{1.000}  &  0.989  /  0.989  &  \textbf{1.000}  /  \textbf{1.000} 
\tabularnewline
winequality red 4 &  0.127  /  0.094  &  0.099  /  0.073  &  0.066  /  0.054  &  \textbf{0.130}  /  \textbf{0.104}  &  0.128  /  0.086  &  0.107  /  0.066 
\tabularnewline
yeast4 &  \textbf{0.377}  /  \textbf{0.349}  &  0.371  /  \textbf{0.349}  &  0.275  /  0.260  &  0.246  /  0.218  &  0.308  /  0.293  &  0.351  /  0.329 
\tabularnewline
yeast5 &  0.714  /  0.705  &  0.690  /  0.680  &  \textbf{0.746}  /  \textbf{0.739}  &  0.617  /  0.603  &  0.716  /  0.707  &  0.724  /  0.716 
\tabularnewline
yeast6 &  0.473  /  0.458  &  \textbf{0.569}  /  \textbf{0.559}  &  0.494  /  0.484  &  0.432  /  0.417  &  0.526  /  0.517  &  0.564  /  0.554 
\tabularnewline
\hline Average &  \textbf{0.656}  /  0.641  &  0.654  /  0.642  &  0.638  /  0.630  &  0.590  /  0.573  &  0.644  /  0.634  &  0.654  /  \textbf{0.642} 
\tabularnewline
\hline\end{tabular}\end{table*}

\begin{table*}[htbp]\scriptsize\caption{Table showing F1-score/$\kappa$- score for compared oversampling strategies for kNN classifier on 14 benchmarked imbalanced datasets. Models with the best performances are marked in bold for each dataset. Convex space modeling approaches ProWRAS and ConvGeN yield a better average performance compared to existing deep-generative models.}\label{tab:results:KNN:A}\centering\tabularnewline\begin{tabular}{l|@{\hskip3pt}c@{\hskip3pt}|@{\hskip3pt}c@{\hskip3pt}|@{\hskip3pt}c@{\hskip3pt}|@{\hskip3pt}c@{\hskip3pt}|@{\hskip3pt}c@{\hskip3pt}|@{\hskip3pt}c@{\hskip3pt}}\hline
\textbf{dataset ($f_1~$score$~/~\kappa~$score)} & \textbf{Repeater} & \textbf{ProWRAS} & \textbf{GAN} & \textbf{CTGAN} & \textbf{CTAB GAN} & \textbf{ConvGeN(min,maj)}
\tabularnewline
\hline
abalone 17 vs 7 8 9 10 &  0.315  /  0.291  &  \textbf{0.336}  /  \textbf{0.313}  &  0.272  /  0.243  &  0.180  /  0.173  &  0.280  /  0.268  &  0.315  /  0.296 
\tabularnewline
abalone9 18 &  0.342  /  0.286  &  \textbf{0.386}  /  \textbf{0.338}  &  0.258  /  0.197  &  0.276  /  0.256  &  0.338  /  0.313  &  0.136  /  0.120 
\tabularnewline
car good &  0.357  /  0.312  &  \textbf{0.720}  /  \textbf{0.707}  &  0.427  /  0.394  &  0.229  /  0.171  &  0.433  /  0.395  &  0.509  /  0.479 
\tabularnewline
car vgood &  0.413  /  0.375  &  0.805  /  0.797  &  0.653  /  0.634  &  0.322  /  0.277  &  0.595  /  0.572  &  \textbf{0.879}  /  \textbf{0.874} 
\tabularnewline
flare F &  0.245  /  0.191  &  0.293  /  0.249  &  0.270  /  0.228  &  0.267  /  0.214  &  0.293  /  0.246  &  \textbf{0.306}  /  \textbf{0.265} 
\tabularnewline
hypothyroid &  0.568  /  0.540  &  0.615  /  0.592  &  0.573  /  0.547  &  \textbf{0.618}  /  \textbf{0.595}  &  0.565  /  0.537  &  0.606  /  0.582 
\tabularnewline
kddcup guess passwd vs satan &  \textbf{0.991}  /  \textbf{0.991}  &  \textbf{0.991}  /  \textbf{0.991}  &  \textbf{0.991}  /  \textbf{0.991}  &  \textbf{0.991}  /  \textbf{0.991}  &  \textbf{0.991}  /  \textbf{0.991}  &  \textbf{0.991}  /  \textbf{0.991} 
\tabularnewline
kr vs k three vs eleven &  0.901  /  0.898  &  0.932  /  0.929  &  0.903  /  0.900  &  0.804  /  0.797  &  0.807  /  0.801  &  \textbf{0.961}  /  \textbf{0.960} 
\tabularnewline
kr vs k zero one vs draw &  0.840  /  0.833  &  0.885  /  0.880  &  0.828  /  0.821  &  0.767  /  0.756  &  0.786  /  0.776  &  \textbf{0.897}  /  \textbf{0.892} 
\tabularnewline
shuttle 2 vs 5 &  0.951  /  0.951  &  \textbf{0.956}  /  \textbf{0.955}  &  0.936  /  0.935  &  0.951  /  0.951  &  0.949  /  0.948  &  0.932  /  0.931 
\tabularnewline
winequality red 4 &  0.069  /  0.016  &  0.072  /  0.020  &  \textbf{0.085}  /  \textbf{0.028}  &  0.047  /  0.001  &  0.057  /  0.003  &  0.058  /  -0.001 
\tabularnewline
yeast4 &  0.329  /  0.293  &  \textbf{0.363}  /  \textbf{0.332}  &  0.332  /  0.294  &  0.304  /  0.267  &  0.313  /  0.273  &  0.348  /  0.313 
\tabularnewline
yeast5 &  0.670  /  0.657  &  \textbf{0.691}  /  \textbf{0.679}  &  0.598  /  0.581  &  0.542  /  0.522  &  0.507  /  0.484  &  0.646  /  0.631 
\tabularnewline
yeast6 &  0.409  /  0.387  &  \textbf{0.463}  /  \textbf{0.445}  &  0.363  /  0.339  &  0.383  /  0.360  &  0.304  /  0.276  &  0.426  /  0.406 
\tabularnewline
\hline Average &  0.529  /  0.502  &  \textbf{0.608}  /  \textbf{0.588}  &  0.535  /  0.509  &  0.477  /  0.452  &  0.516  /  0.492  &  0.572  /  0.553 
\tabularnewline
\hline\end{tabular}\end{table*}

\section{Results}\label{sec: results}

\begin{figure*}[ht]

\centering
\includegraphics[width=0.49\textwidth]{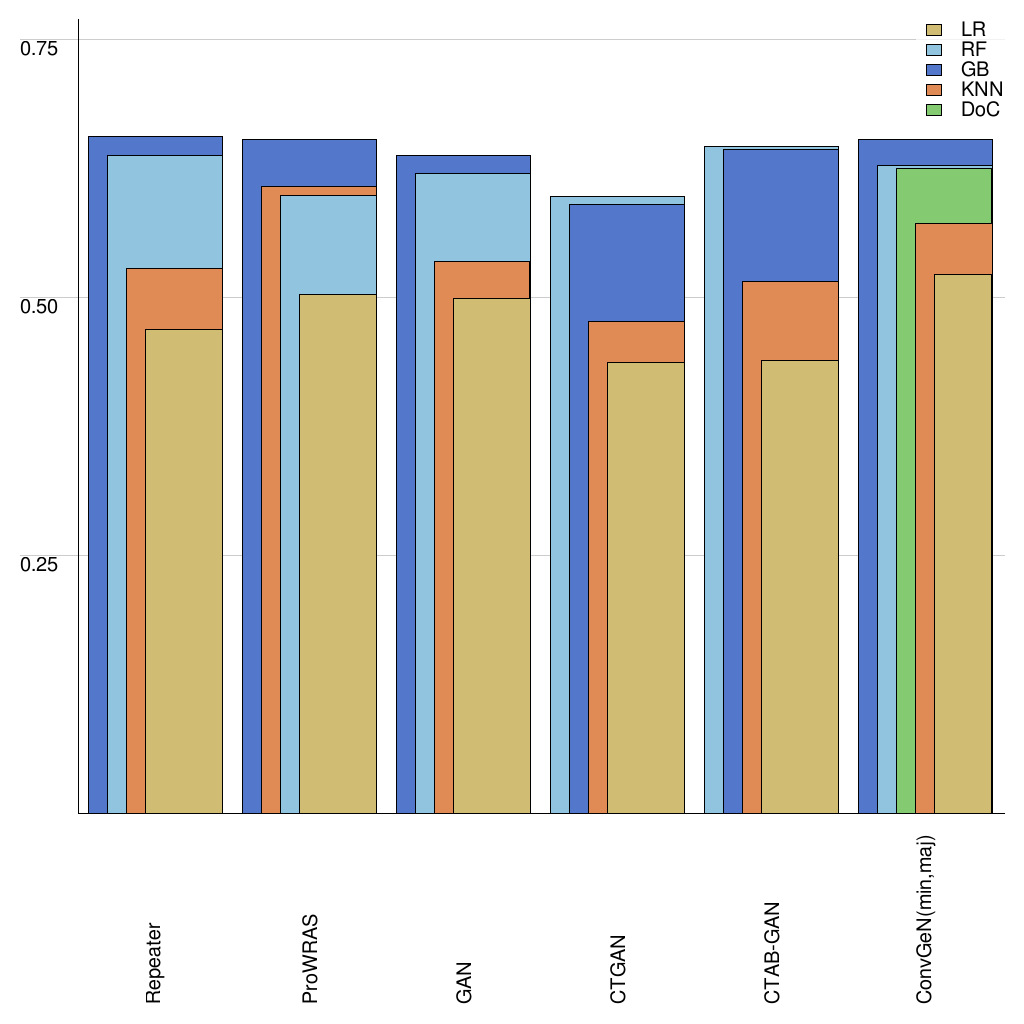} 
\hfill
\includegraphics[width=0.49\textwidth]{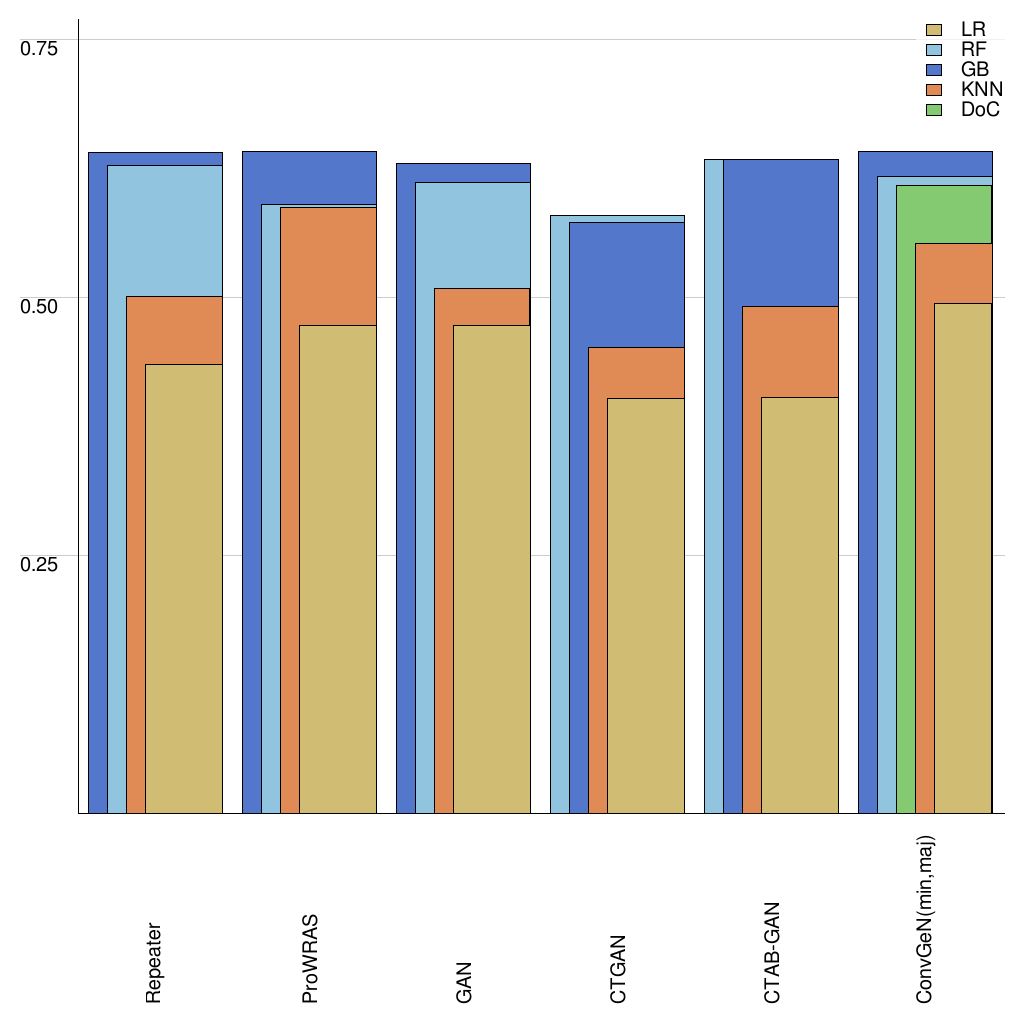} 

\caption{Average by oversampling algorithms for $f_1~$score (on the left) and $\kappa~$score (on the right) for different classifiers over 14 benchmarking datasets. Ensemble models yield better average classification performance.}
\label{fig:results:classifier}
\end{figure*}

\begin{figure*}[!htbp]

\centering
\includegraphics[width=0.49\textwidth]{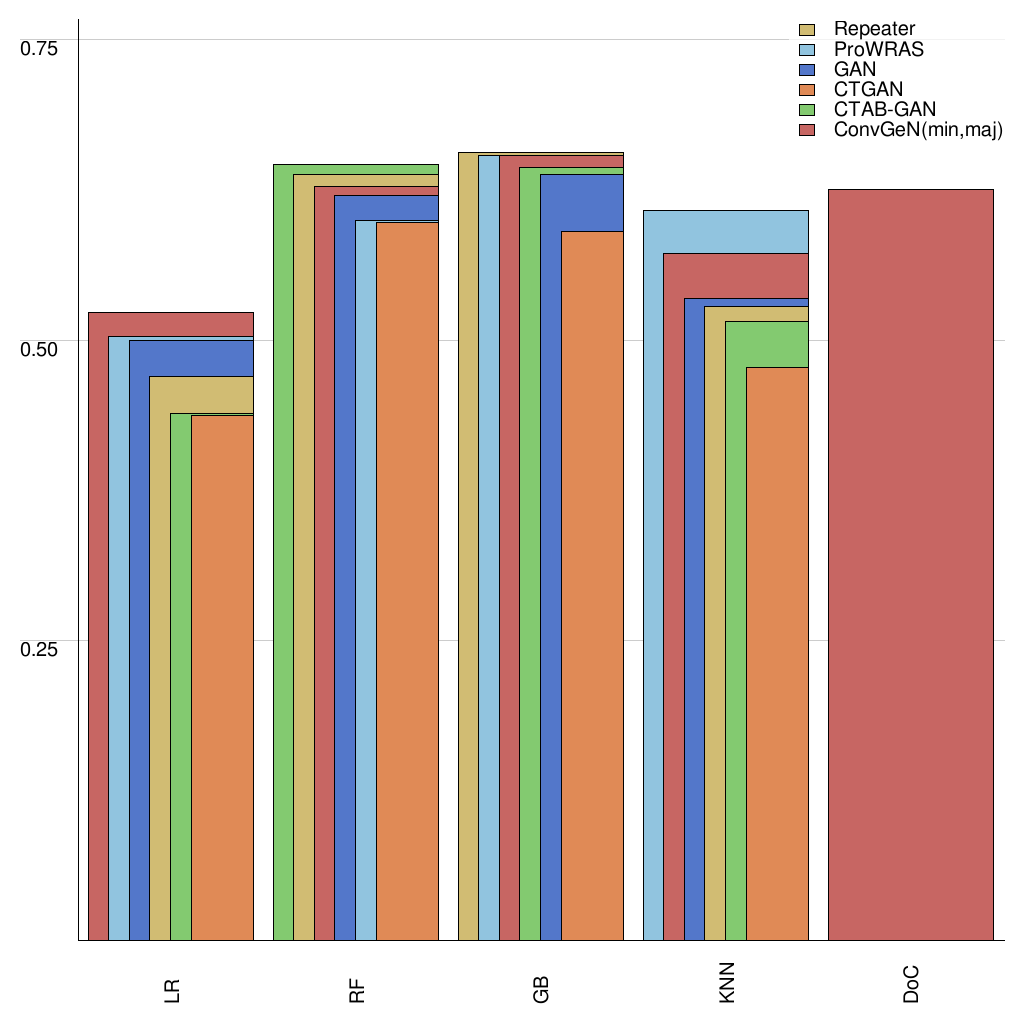} 
\hfill
\includegraphics[width=0.49\textwidth]{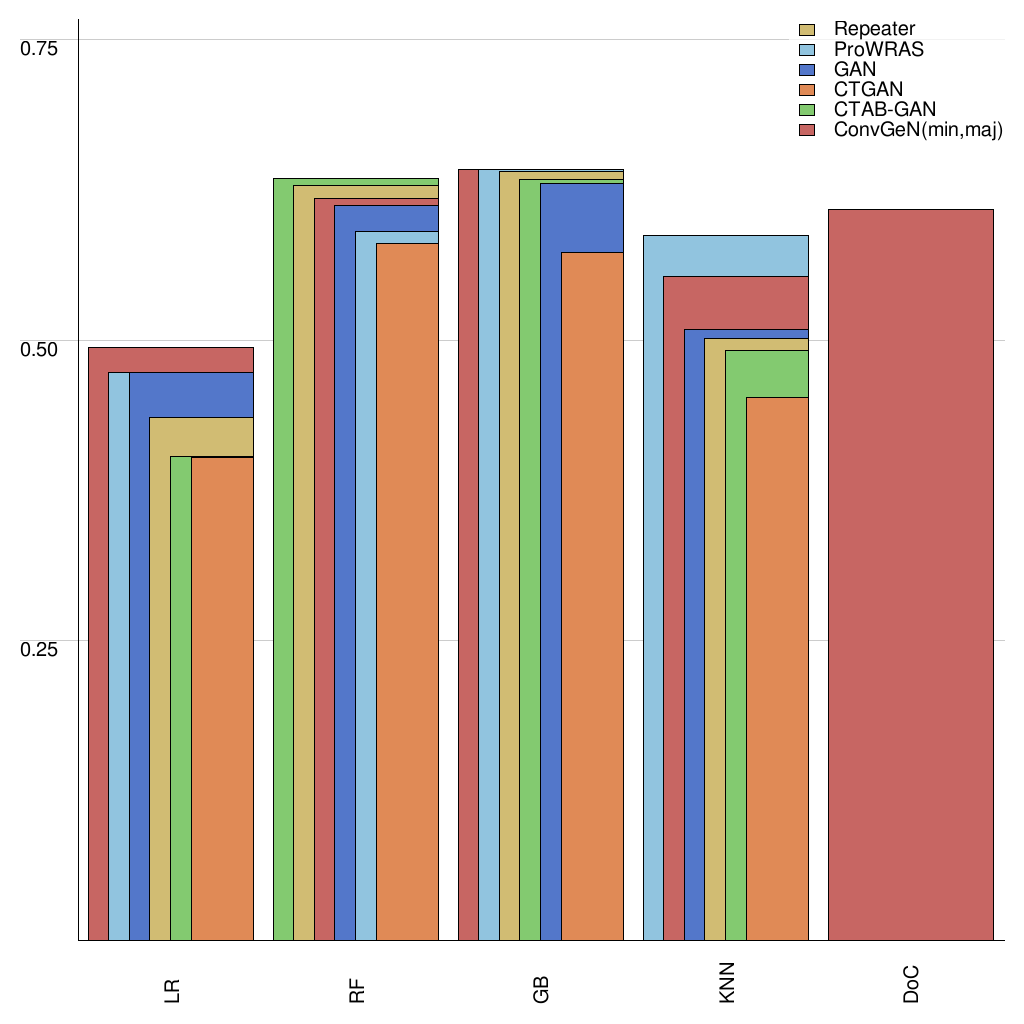} 

\caption{Average by classifier for $f_1~$score (on the left) and $\kappa~$score (on the right) for different oversampling models over 14 benchmarking datasets. Note that the ConvGeN(min,maj) model performs either the best or the second best for all classifiers. For every classifier, it performs better than the existing deep-generative models.}
\label{fig:results:algo}
\end{figure*}

\textbf{Ensemble models perform better on smaller tabular datasets.} (See Figure \ref{fig:results:classifier}) Our benchmarking studies indicate that ensemble models GB and RF perform better in terms of both the F1-score and $\kappa$-score, for smaller tabular datasets. Such models rely on an ensemble of weak classifiers for decision-making. However, we would like to point out here that all our benchmarking datasets contain a relatively low number of features (see Table \ref{tbl:methods:datasets}). Such a scenario is ideal for ensemble models. For high-dimensional datasets, such models tend to be less robust. In such a scenario, simpler models are more reliable \cite{overfitting_RF}. We also noticed that the kNN is consistently better than the LR. Moreover, for the ProRWAS oversampling algorithm, the kNN performs even better than the RF model contrary to Bej \textit{et al.}, as we did not customize ProWRAS for RF, that is to choose the proper oversampling scheme for every dataset \cite{ProWRAS}. The performance of the DoC classifier is highly comparable to the RF model. However, we notice that, it performs better than the kNN and LR classifier for all oversampling algorithms. \par

\textbf{The baseline Repeater model is compatible with ensemble models.} (See Figure \ref{fig:results:algo}) An interesting phenomenon that we inferred from our results is that the baseline repeater model performs the second best for RF in terms of both F1-Score and $\kappa$-Score. For GB, it performs best in terms of F1-Score and third best in terms of $\kappa$-Score. This shows that for ensemble models on datasets with fewer features, there might be not much need to use complex oversampling algorithms. Rather, balancing the classes by simply bootstrapping on the minority class can be good enough. However, for simpler models, such as kNN and LR, which might be more effective in specific scenarios, we observe that this approach does not improve the classification performance. Synthetic sample generation using convex space learning is clearly the most effective strategy of data rebalancing in such cases.\par

\textbf{ConvGeN performs better than existing state-of-the-art deep-generative models for tabular datasets in the context of imbalanced classification.} (See Figure \ref{fig:results:algo})
It is evident from our experiments that ConvGeN(min,maj) outperforms the conventional GAN model, as well as the CTGAN models for smaller tabular imbalanced datasets for every classifier. CTAB-GAN performs better than ConvGeN only for the RF classifier. Interestingly, for smaller tables, we observed that conventional GAN works better than CTGAN for all classifiers. GAN also performs better than CTAB-GAN for all non-ensemble classifiers (LR and kNN). On the other hand, among the two GAN variants, CTAB-GAN performs better than CTGAN consistently for all classifiers. \par


\textbf{Convex space learning approaches are better for simpler models like kNN and LR models.} (See Figure \ref{fig:results:algo})
Convex space learning approaches are better for simpler models like kNN and LR models. For the kNN algorithm however, the ProWRAS approach clearly outperforms ConvGeN(min,maj), while ConvGeN(min,maj) still performs better than the other GAN based approaches. For the LR algorithm, the ConvGeN(min,maj) approach clearly outperforms ProWRAS, while ProWRAS still performs better than the other GAN based approaches. Since both ProWRAS and ConvGeN(min,maj) rely on the philosophy of convex space learning, we can conclude that this approach is better suited for simpler classifiers, such as kNN and LR, compared to existing deep-generative approaches.\par

\textbf{Convex space learning is more suitable for oversampling on smaller tabular imbalanced datasets:} (See table \ref{tab:rank})
We have ranked the oversampling models according to their performance the classifiers GB, RF, kNN, LR based on F1 and $\kappa$ scores. We observe that ConvGeN produces the best average rank for both the performance measures, while ProWRAS produces the second-best average rank. Interestingly these are the only two models that produce an average rank better than the baseline Repeater model. Thus, we can conclude that convex space learning is a more suitable synthetic sample generation strategy compared to deep adversarial learning, when the synthetic samples are to be used for class rebalancing, especially for smaller datasets.

\begin{table}[!ht]
\caption{Oversampling models are ranked according to their performance the classifiers GB, RF, kNN, LR based on F1 and $\kappa$ scores. ConvGen(min,maj) produces the best average rank over all classifiers and ProWRAS produces the second best performance. }\label{tab:rank}
    \centering
    \resizebox{\columnwidth}{!}{\begin{tabular}{l|@{\hskip3pt}c@{\hskip3pt}|@{\hskip3pt}c@{\hskip3pt}|@{\hskip3pt}c@{\hskip3pt}|@{\hskip3pt}c@{\hskip3pt}|@{\hskip3pt}c@{\hskip3pt}|@{\hskip3pt}c@{\hskip3pt}}
        \hline
         & \textbf{Repeater} & \textbf{ProWRAS} & \textbf{GAN} & \textbf{CTGAN} & \textbf{CTAB-GAN} & \textbf{ConvGen(min,maj)}
        \tabularnewline
        \hline
        GB (F1-score) & 1 & 2 & 5 & 6 & 4 & 3
        \tabularnewline
        RF (F1-score) & 2 & 5 & 4 & 6 & 1 & 3 
        \tabularnewline
        kNN (F1-score) & 4 & 1 & 3 & 6 & 5 & 2
        \tabularnewline
        LR (F1-score) & 4 & 2 & 3 & 6 & 5 & 1
        \tabularnewline
        \hline
        \textbf{Average} rank (F1-score) & 2.75 & 2.5 & 3.75 & 6 & 3.75 & \textbf{2.25}
        \tabularnewline
        \hline
        GB ($\kappa$-score) & 3 & 2 & 5 & 6 & 4 & 1
        \tabularnewline
        RF ($\kappa$-score) & 2 & 5 & 4 & 6 & 1 & 3
        \tabularnewline
        kNN ($\kappa$-score) & 4 & 1 & 3 & 6 & 5 & 2
        \tabularnewline
        LR ($\kappa$-score) & 4 & 2 & 3 & 6 & 5 & 1
        \tabularnewline
        \hline
        \textbf{Average rank ($\kappa$-score)} & 3.25 & 2.5 & 3.75 & 6 & 3.75 & \textbf{1.75}
        \tabularnewline \hline
    \end{tabular}}
\end{table}

\begin{table}[!ht]
\caption{The p-values for a Wilcoxon's signed rank test comparing ConvGeN, with each of the benchmarked oversampling models, and performance metric, pairwise. Case studies for which ConvGeN improves significantly over a compared model (p-value$<0.05$) are marked in bold font.}\label{tab:Wilcox}
    \centering
    \resizebox{\columnwidth}{!}{\begin{tabular}{l|@{\hskip3pt}c@{\hskip3pt}|@{\hskip3pt}c@{\hskip3pt}|@{\hskip3pt}c@{\hskip3pt}|@{\hskip3pt}c@{\hskip3pt}|@{\hskip3pt}c@{\hskip3pt}}
        \hline
         & \textbf{Repeater} & \textbf{ProWRAS} & \textbf{GAN} & \textbf{CTGAN} & \textbf{CTAB-GAN} 
        \tabularnewline
        \hline
        GB (F1-score) & 0.85 & 0.62 & 0.23 & 0.05 & 0.38 
        \tabularnewline
        GB ($\kappa$-score)& 1 & 0.85 & 0.30 & 0.04 & 0.43
        \tabularnewline
        RF (F1-score) & 0.92 & \textbf{0.02} & 0.27 & 0.20 & 0.23 
        \tabularnewline
        RF ($\kappa$-score) & 1 & 0.01 & 0.28 & 0.11 & 0.34
        \tabularnewline
        kNN (F1-score) & 0.13 & 0.13 & \textbf{0.03} & \textbf{0.03} & \textbf{0.03}  
        \tabularnewline
        kNN ($\kappa$-score) & 0.12 & 0.13 & \textbf{0.03} & \textbf{0.03} & \textbf{0.03} 
        \tabularnewline
        LR (F1-score) & \textbf{0.00} & \textbf{0.03} & 0.15 & \textbf{0.00} & \textbf{0.00}
        \tabularnewline
        LR ($\kappa$-score) & \textbf{0.00} & 0.05 & 0.24 & \textbf{0.00} & \textbf{0.00}
        \tabularnewline \hline
    \end{tabular}}
\end{table}

\begin{table*}[ht]\scriptsize\caption{Table showing F1-score/$\kappa$-score comparisons among different parameter settings for ConvGeN. The ConvGeN(min,maj) model is  seen to perform the best on average for all the tested classifiers. }\label{tab:results:DOC}\centering\tabularnewline\begin{tabular}{l|@{\hskip3pt}c@{\hskip3pt}|@{\hskip3pt}c@{\hskip3pt}|@{\hskip3pt}c@{\hskip3pt}|@{\hskip3pt}c@{\hskip3pt}}\hline
\textbf{dataset ($f_1~$score$~/~\kappa~$score)} & \textbf{ConvGeN(5,maj)} & \textbf{ConvGeN(min,maj)} & \textbf{ConvGeN(5,prox)} & \textbf{ConvGeN(min,prox)}
\tabularnewline
\hline
LR-Average &  0.489  /  0.457  &  \textbf{0.523}  /  \textbf{0.494}  &  0.488  /  0.457  &  0.515  /  0.486 
\tabularnewline
 RF-Average &  0.600  /  0.591  &  \textbf{0.628}  /  \textbf{0.618}  &  0.604  /  0.594  &  0.622  /  0.611 
\tabularnewline
GB-Average &  0.627  /  0.615  &  \textbf{0.654}  /  \textbf{0.642}  &  0.631  /  0.619  &  0.648  /  0.637 
\tabularnewline
kNN-Average &  0.557  /  0.532  &  \textbf{0.572}  /  \textbf{0.553}  &  0.557  /  0.532  &  0.569  /  0.550 
\tabularnewline
DoC-Average & 0.611  /  0.590  &  \textbf{0.625}  /  \textbf{0.609}  &  0.614  /  0.594  &  0.614  /  0.596 
\tabularnewline
\hline
\end{tabular}\end{table*}

Following the work of Bej \textit{et al.}, we also tested the performance of the ConvGeN(min,maj) model against other oversampling models using Wilcoxon's signed rank test\cite{ProWRAS}. We used the {\fontfamily{pcr}\selectfont scipy.stats} library for this test. A limitation of the library is that exact p-value calculation does not work if there are ties. Since there are ties in our results, the test produces approximate measures. The results of the test are presented in Table \ref{tab:Wilcox}. We observe that ConvGeN(min,maj) significantly improves model performance for non ensemble models, such as kNN and LR, compared to existing deep-generative models, especially CTGAN and CTAB-GAN, confirming that convex space modeling is indeed more suitable for tabular imbalanced datasets of small size compared to deep adversarial learning.\par

\textbf{ConvGeN(min,maj) model produces the best classification scores among different ConvGeN parameter settings.}
We tested four different parameter settings for the ConvGeN model. We have provided in Table \ref{tab:results:DOC} a comparison of the ConvGeN model with these four parameter settings. For all the classifiers that we have tested, including the DoC classifier, we observe that ConvGeN(min,maj) performs the best. These parameter settings for ConvGeN mean that we consider the entire minority class as the minority neighborhood and use no constraint on the majority class. However, it is important to note that all datasets under investigation for benchmarking have a small minority class. That is why the parameter \var{neb} can be an important addition to the model for datasets dealing with larger data with possibly a larger minority class. Interestingly, we observe that among the models ConvGeN(5,maj) and ConvGeN(5,prox), the latter works better. From this, we conclude that restricting sampling majority batches from a proximal majority subset can be useful when we are modeling smaller minority neighborhoods, that is when we are using a smaller value for the parameter \var{neb}. This indicates that both these parameters could be useful in improving the applicability of the model.\par

\section{Discussion}\label{sec: discussion}

\begin{figure*}[!htbp]
\label{fig:Discussion:comparison}
\centering
\includegraphics[width=\textwidth]{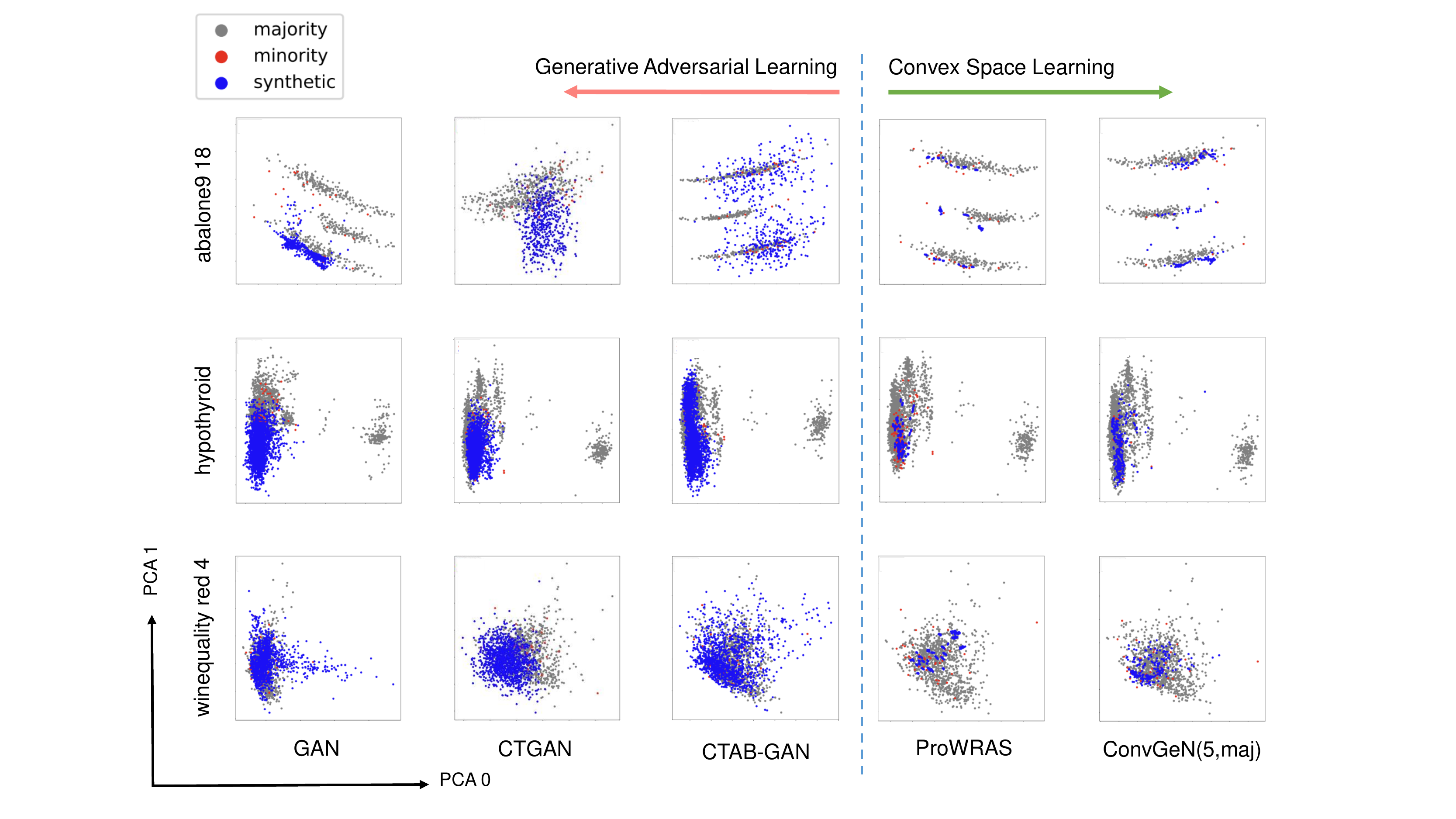} 
\caption{Figure showing synthetic sample generation for different models on three datasets with diverse data distributions. We observe that approaches based on the philosophy of convex space modeling, especially ProWRAS and ConvGeN(5,maj) generate synthetic data that are visually closer to the minority data space.}

\end{figure*}

First, we point out some limitations of the ConvGeN model. The model does not include a mechanism for categorical data generation in particular. This might limit its applicability for synthetic data generation of tabular data, in general. Additionally, the model has not been tested for multi-class classification. Nevertheless, the model can certainly be adapted for multi-classification, simply by choosing the discriminator to be multi-class and training the generator for all minority classes. However, well-established multi-class imbalanced datasets for benchmarking were not found in the literature from related works. Moreover, we also did not experiment extensively with the complexity and architecture of the model, such as increasing the depth of the neural networks, use of activation functions, regularization, loss functions, and optimizers. This was due to the fact that the benchmarking datasets were mostly low dimensional and therefore, using simple architectures would already provide us with a good enough idea about the effect of convex space learning on tabular imbalanced classification on small datasets.\par

In this article, we introduced the ConvGeN model for synthetic data generation particularly in the context of imbalanced classification. However, the scope of synthetic sample generation can also be set on a larger scale. In particular, in clinical data science, synthetic data can be used to likewise build and optimize machine learning models and protect the privacy of patients \cite{privacy}. Note that the ConvGeN model can be easily adapted to generate synthetic data from entire datasets. In such a case, we can train the generator using neighborhood batches from entire datasets, and the discriminator can be trained to differentiate a data neighborhood from a randomly sampled shuffled batch from the complement of the neighborhood, which is passed to the generator during one training iteration. The distribution of the real data can be considered to decide on the distribution of the synthetic data. However, the current version of ConvGeN does not have a particular strategy for generating categorical data in particular, which would be crucial to broaden its applicability for future applications. Unfortunately, the benchmarking datasets commonly available for imbalanced classification mostly lack diverse categorical data types, for which the applicability of ConvGeN could not be investigated. A more general version of the ConvGeN algorithm can find applications in several research domains, such as clinical and finance. Such a model could help circumvent real world problems like prevention of data sharing due to privacy issues and class imbalance in clinical trials \cite{privacy, privacy2} and as byproducts can generate necessary benchmarking datasets for categorical tabular imbalanced classification.  However, some crucial observations arise from this research, which are discussed below.\par
Deep-generative models, since their inception, have received a lot of attention from data scientists. A vast amount of research has been done on GANs in the domain of synthetic image generation. The capability of GANs to create realistic images have been proven time and again. However, for image datasets, we have a perceptional advantage, in the sense, that we can visually judge how realistic the synthetic image is, just by looking at it. In particular, for a medical image, a professional radiologist can easily verify whether a GAN-generated image looks realistic enough. This method of validation, popularly used in clinical sciences is called a Visual Turing test \cite{Turing_test}. Unfortunately for tabular datasets, we lose the advantage of clear visual perception and there is no known analog for a Visual Turing test for tabular data. That is why for such datasets, it is difficult to judge how realistic the generated data is. The best we can do is to construct a two or three dimensional projection, using semi-supervised algorithms.\par
In Figure \ref{fig:Discussion:comparison}, we show Principal component analysis (PCA) plots of real and synthetic data generated by the oversampling models: ProWRAS, GAN, CTAB-GAN, CTGAN, and ConvGeN(5,maj) for three datasets from our benchmarking study: `abalone9 18', `hypothyroid', and `winequality red 4'. Note that the 'abalone9 18' dataset has three clusters, each of which have some minority samples, the `hypothyroid' dataset has two clusters where the minority samples are present in only one cluster and there are no particularly visible clusters in the `winequality red 4' dataset. We can observe that the data generated by ProWRAS and ConvGeN(5,maj) are well-embedded in the minority data space and hence can be interpreted visually as realistic compared to the rest.  \par
Our experiments also raise the question, whether synthetic sample generation should be task-specific, that is, whether different approaches/algorithms of synthetic data generation are more effective/applicable for diverse machine-learning tasks, say unsupervised data stratification or classification. A common perception might be that the more realistic the synthetic samples, the better the classification. However, it is clear from Figure \ref{fig:Discussion:comparison}, that this is not necessarily the case. For example, for the hypothyroid dataset, the RF model along with the generation of synthetic samples using CTAB-GAN is the best among all models. But from Figure \ref{fig:Discussion:comparison}, visual perception at least indicates that the synthetic data generated by CTAB-GAN for this dataset has the highest variance or spread among all the models. Our results can be considered as a strong evidence that convex space modeling does improve classifier performances compared to existing deep-generative models, such as GAN, CTGAN, and CTAB-GAN for smaller imbalanced tabular datasets. Moreover, the synthetic sample distribution is visually more consistent with the original data distribution.\par

\section{Conclusion}
Our conducted research revealed significant findings in the context of imbalanced classification for which ensemble models have been so far known to perform better. Linear interpolation-based models that generate synthetic samples from the convex space of the minority class outperform nonlinear deep-generative models for small-size tabular imbalanced datasets. We also conclude that visually realistic synthetic data generation (based on PCA plots) does not necessarily guarantee good classification performance for such datasets. Thereby, we point out the scope of research for task-specific synthetic data generation. ConvGeN improves imbalanced classification as compared to existing deep-generative models, such as GAN, CTGAN, and CTAB-GAN for tabular imbalanced datasets of small size. In conclusion, we assert that convex space learning has a broad applicative potential, even outside the context of imbalanced classification, since the synthetic sample distribution is visually more consistent with the original data distribution, in addition to improving imbalanced classification. 

\section*{Author Contribution}

SB proposed the basic architecture of ConvGeN. KS experimented with the idea extensively to improve it further and developed the final ConvGeN architecture. KS worked on the computational implementation and benchmarking studies. KS and SB wrote the first version of the manuscript and are the joint first authors of this work. WH, MW, PS, and OW critically discussed and revised the approach and content of the manuscript. All the authors have read and agreed to the contents of the manuscript.

\section*{Conflict of Interest}
The authors have no conflict of interest.

\section*{Availability of code and results}
To support transparency, re-usability, and reproducibility, we provide an implementation of the algorithm for binary classification problems using {\fontfamily{pcr}\selectfont  Python (V 3.8.10)} and several {\fontfamily{pcr}\selectfont Jupyter Notebooks} from our benchmarking study in a \href{https://github.com/kristian10007/ConvGeN}{ConvGeN GitHub repository}. Other important python library versions we used are {\fontfamily{pcr}\selectfont  keras (V 2.8.0)}, {\fontfamily{pcr}\selectfont  torch (V 1.11.0)}, {\fontfamily{pcr}\selectfont  tensorflow (V 2.8.0)}, {\fontfamily{pcr}\selectfont  scikit-learn (V 1.0.2)}, {\fontfamily{pcr}\selectfont  numpy (V 1.22.3)} and {\fontfamily{pcr}\selectfont  ctgan (V 0.5.1)}.

\section*{Acknowledgment}
We thank the German Network for Bioinformatics Infrastructure (de.NBI) and  Establishment of Systems Medicine Consortium in Germany e:Med for their support, as well as the German Federal Ministry for Education and Research (BMBF) program (FKZ 01ZX1709C) and the EU Social Fund (ESF/14-BM-A55-0027/18) for funding.

\clearpage
\newpage

\begin{IEEEbiography}[{\includegraphics[width=1in,height=1.25in,clip,keepaspectratio]{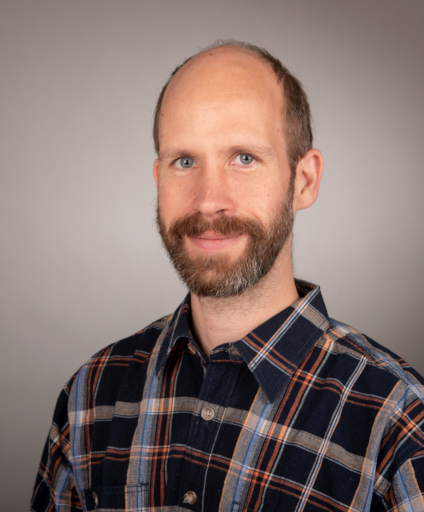}}]{Kristian Schultz} received his first degrees in mathematics and computer science in 2014. After that, he worked for two years in the field of discrete mathematics on Sperner families in the Deptartment of Mathematics at the University in Rostock, Germany. In between, he extended his software developing skills in industry. Since 2020 he returned to the University of Rostock, Department of Systems Biology and Bioinformatics, where he focuses on the correctness, efficiency, and implementation of algorithms. 
\end{IEEEbiography}

\begin{IEEEbiography}[{\includegraphics[width=1in,height=1.25in,clip,keepaspectratio]{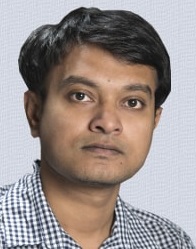}}]{Saptarshi Bej} received his Master’s degree in Mathematics from Indian Institute of Science Education and Research, Kolkata in 2014. In 2021, he completed his PhD in Computer Science from the University of Rostock, Germany. He is currently a research associate at the Institute of Computer Science, University of Rostock, Germany and a guest scientist in the Leibniz-Institute for Food Systems Biology at the Technical University of Munich, Germany. His primary research focus is on the development of machine learning algorithms and their applications in the life sciences. His special interests include machine learning on small and imbalanced datasets and literature mining. 
\end{IEEEbiography}

\begin{IEEEbiography}[{\includegraphics[width=1in,height=1.25in,clip,keepaspectratio]{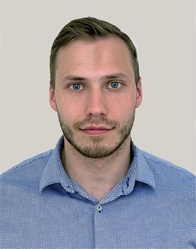}}]{Waldemar Hahn} received his Master’s degree in Computer Science in 2019. Since 2020 he works as a research assistant at the Center for Medical Informatics within the Institute for Medical Informatics and Biometry (Faculty of Medicine Carl Gustav Carus, Technische Universität Dresden). His current research is related to the generation and evaluation of tabular data in healthcare.
\end{IEEEbiography}

\begin{IEEEbiography}[{\includegraphics[width=1in,height=1.25in,clip,keepaspectratio]{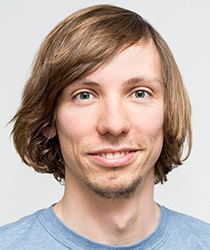}}]{Markus Wolfien} received his Master’s degree in Medical Biotechnology and completed his PhD on the topic `Customized workflow development and omics data integration concepts in systems medicine' from the University of Rostock. He is currently leading the ``Data Science'' research group at the Center for Medical Informatics within the Institute for Medical Informatics and Biometry (Faculty of Medicine Carl Gustav Carus, Technische Universität Dresden). He focuses on the application of AI-based approaches and develops flexible workflows for the joint analysis of low and high-throughput data, such as clinical blood measurements, protein expression, RNA sequencing data (bulk, single cell, and spatial), and environmental information. To achieve this, he combines state-of-the-art tools including R and Python, as well as further downstream analysis approaches, such as network analyses and machine learning in general (classical ML and Deep Learning). 
\end{IEEEbiography}

\begin{IEEEbiography}[{\includegraphics[width=1in,height=1.25in,clip,keepaspectratio]{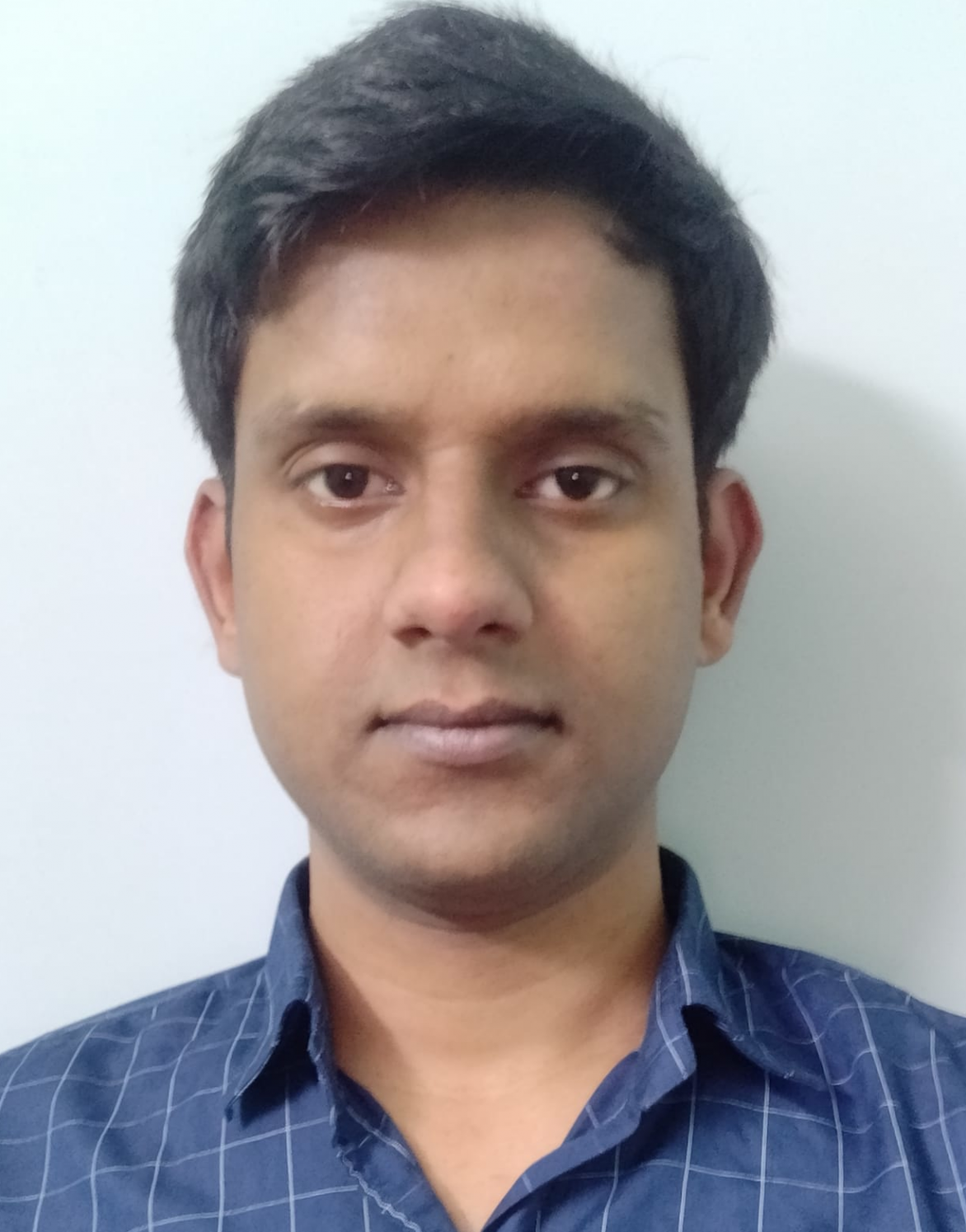}}]{Prashant Srivastava} received his Master’s degree in Physics from Indian Institute of Science Education and Research, Kolkata. After that he did a second Master's degree in Data Science and Analytics from Department of Computer Science, Royal Holloway, University of London. He is currently a research assistant in the Department of Systems Biology and Bioinformatics, Institute of Computer Science at the University of Rostock, Germany. His current research interests are algorithm development and their applications in literature mining. 
\end{IEEEbiography}

\begin{IEEEbiography}[{\includegraphics[width=1in,height=1.25in,clip,keepaspectratio]{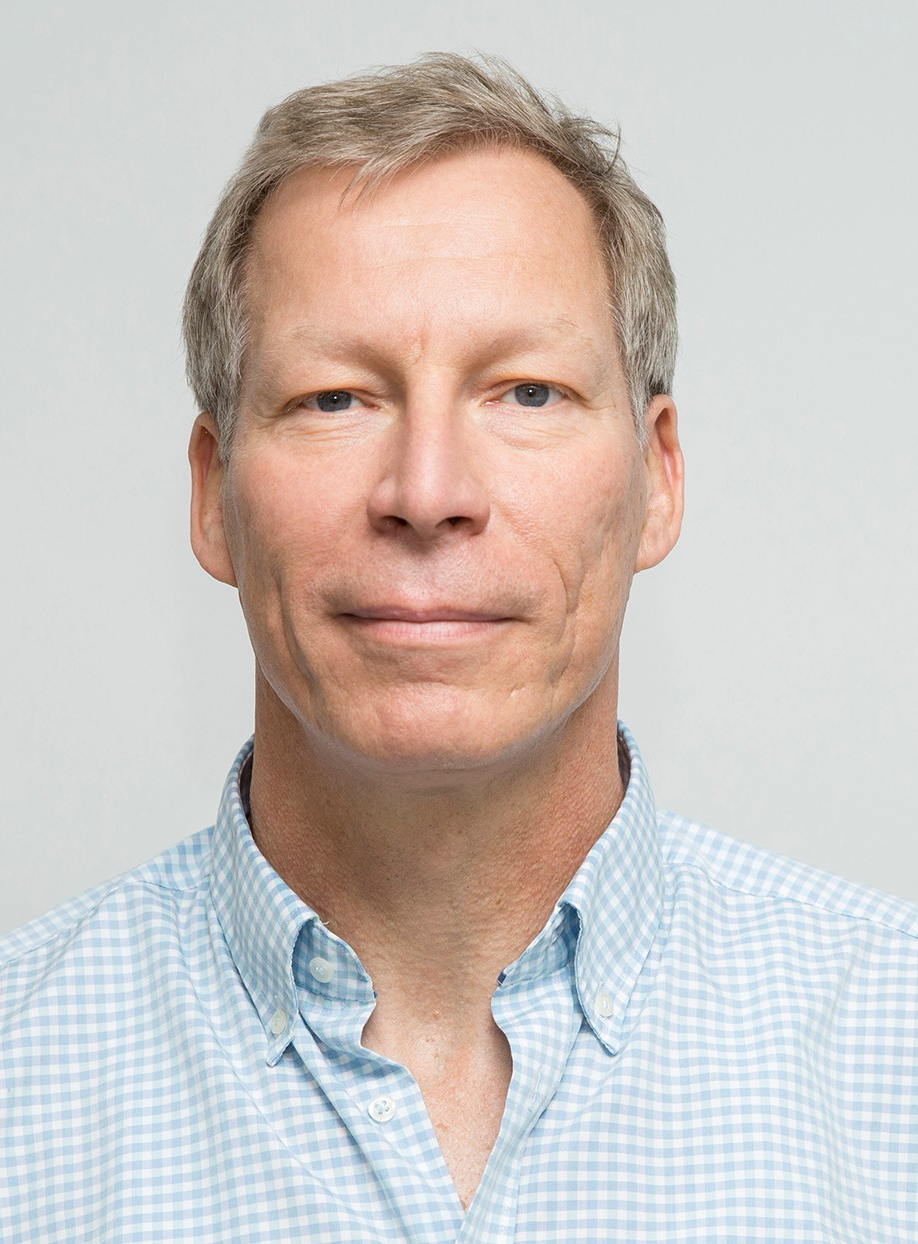}}]{Olaf Wolkenhauer}  received his first degrees in systems and control engineering and his PhD for research in possibility theory with applications to data analysis. He spent over ten years at the University of Manchester Institute of Science and Technology in the UK. In 2005, he became a fellow of the Stellenbosch Institute for Advanced Study and holds professorships at Case Western Reserve University, USA and Chhattisgarh Swami Vivekanand Technical University, India. In 2003, he was appointed as professor for systems biology and bioinformatics at the University of Rostock, Germany. In 2015, he was elected a member of the Foundations in Medicine and Biology review panel of the German Research Foundation (DFG). His research combines data-driven modelling with model-driven experimentation, using a wide range of approaches, including machine learning, and systems theory.
\end{IEEEbiography}

\vfill
\clearpage
\newpage

\section{Supplementary data}\label{suppl}

 \begin{table*}[ht]\scriptsize\caption{Table showing F1-score/$\kappa$-score for different parameter settings for ConvGeN, using the LR classifier. The ConvGeN(min,maj) model is  seen to perform the best for all the tested classifiers.}\label{tab:results:LR:B}\centering\tabularnewline\begin{tabular}{l|@{\hskip3pt}c@{\hskip3pt}|@{\hskip3pt}c@{\hskip3pt}|@{\hskip3pt}c@{\hskip3pt}|@{\hskip3pt}c@{\hskip3pt}}\hline
\textbf{dataset ($f_1~$score$~/~\kappa~$score)} & \textbf{ConvGeN(5,maj)} & \textbf{ConvGeN(min,maj)} & \textbf{ConvGeN(5,prox)} & \textbf{ConvGeN(min,prox)}
\tabularnewline
\hline
abalone 17 vs 7 8 9 10 &  0.292  /  0.262  &  0.337  /  0.310  &  0.292  /  0.261  &  \textbf{0.344}  /  \textbf{0.318} 
\tabularnewline
abalone9 18 &  0.538  /  0.500  &  \textbf{0.575}  /  \textbf{0.544}  &  0.541  /  0.503  &  0.574  /  0.543 
\tabularnewline
car good &  0.108  /  0.038  &  0.109  /  0.038  &  0.107  /  0.036  &  0.110  /  0.039 
\tabularnewline
car vgood &  0.373  /  0.332  &  \textbf{0.422}  /  \textbf{0.387}  &  0.377  /  0.337  &  0.400  /  0.362 
\tabularnewline
flare F &  0.273  /  0.223  &  0.323  /  0.279  &  0.276  /  0.226  &  0.317  /  0.274 
\tabularnewline
hypothyroid &  0.363  /  0.312  &  0.378  /  0.328  &  0.360  /  0.308  &  \textbf{0.429}  /  \textbf{0.386} 
\tabularnewline
kddcup guess passwd vs satan &  0.989  /  0.989  &  0.995  /  0.995  &  0.994  /  0.994  &  \textbf{0.998}  /  \textbf{0.998} 
\tabularnewline
kr vs k three vs eleven &  0.952  /  0.951  &  \textbf{0.965}  /  \textbf{0.965}  &  0.948  /  0.947  &  0.963  /  0.962 
\tabularnewline
kr vs k zero one vs draw &  0.700  /  0.686  &  \textbf{0.836}  /  \textbf{0.830}  &  0.710  /  0.696  &  0.761  /  0.750 
\tabularnewline
shuttle 2 vs 5 &  0.994  /  0.994  &  0.996  /  0.996  &  0.989  /  0.989  &  0.992  /  0.992 
\tabularnewline
winequality red 4 &  0.130  /  0.075  &  0.135  /  0.080  &  0.126  /  0.071  &  0.131  /  0.076 
\tabularnewline
yeast4 &  0.254  /  0.209  &  0.277  /  0.235  &  0.249  /  0.204  &  0.276  /  0.234 
\tabularnewline
yeast5 &  0.582  /  0.564  &  \textbf{0.620}  /  \textbf{0.604}  &  0.577  /  0.559  &  0.601  /  0.584 
\tabularnewline
yeast6 &  0.295  /  0.266  &  0.353  /  0.328  &  0.290  /  0.261  &  0.316  /  0.289 
\tabularnewline
\hline Average &  0.489  /  0.457  &  \textbf{0.523}  /  \textbf{0.494}  &  0.488  /  0.457  &  0.515  /  0.486 
\tabularnewline
\hline\end{tabular}\end{table*}

\begin{table*}[ht]\scriptsize\caption{Table showing F1-score/$\kappa$-score for different parameter settings for ConvGeN, using the RF classifier. The ConvGeN(min,maj) model is  seen to perform the best for all the tested classifiers.}\label{tab:results:RF:B}\centering\tabularnewline\begin{tabular}{l|@{\hskip3pt}c@{\hskip3pt}|@{\hskip3pt}c@{\hskip3pt}|@{\hskip3pt}c@{\hskip3pt}|@{\hskip3pt}c@{\hskip3pt}}\hline
\textbf{dataset ($f_1~$score$~/~\kappa~$score)} & \textbf{ConvGeN(5,maj)} & \textbf{ConvGeN(min,maj)} & \textbf{ConvGeN(5,prox)} & \textbf{ConvGeN(min,prox)}
\tabularnewline
\hline
abalone 17 vs 7 8 9 10 &  0.234  /  0.222  &  0.280  /  0.268  &  0.253  /  0.240  &  \textbf{0.283}  /  \textbf{0.270} 
\tabularnewline
abalone9 18 &  0.324  /  0.291  &  0.348  /  0.326  &  0.335  /  0.302  &  0.359  /  0.337 
\tabularnewline
car good &  0.634  /  0.624  &  0.659  /  0.650  &  0.655  /  0.646  &  0.647  /  0.638 
\tabularnewline
car vgood &  0.910  /  0.907  &  0.929  /  0.927  &  0.904  /  0.901  &  0.921  /  0.918 
\tabularnewline
flare F &  0.062  /  0.043  &  0.120  /  0.099  &  0.064  /  0.044  &  0.133  /  0.113 
\tabularnewline
hypothyroid &  0.782  /  0.772  &  \textbf{0.798}  /  \textbf{0.788}  &  0.791  /  0.781  &  0.796  /  0.786 
\tabularnewline
kddcup guess passwd vs satan &  \textbf{1.000}  /  \textbf{1.000}  &  \textbf{1.000}  /  \textbf{1.000}  &  \textbf{1.000}  /  \textbf{1.000}  &  \textbf{1.000}  /  \textbf{1.000} 
\tabularnewline
kr vs k three vs eleven &  0.993  /  0.993  &  0.992  /  0.992  &  0.993  /  0.993  &  0.992  /  0.992 
\tabularnewline
kr vs k zero one vs draw &  0.941  /  0.939  &  0.946  /  0.944  &  0.937  /  0.935  &  \textbf{0.949}  /  \textbf{0.947} 
\tabularnewline
shuttle 2 vs 5 &  \textbf{1.000}  /  \textbf{1.000}  &  \textbf{1.000}  /  \textbf{1.000}  &  \textbf{1.000}  /  \textbf{1.000}  &  \textbf{1.000}  /  \textbf{1.000} 
\tabularnewline
winequality red 4 &  0.079  /  0.062  &  0.086  /  0.055  &  0.067  /  0.050  &  0.070  /  0.041 
\tabularnewline
yeast4 &  0.251  /  0.240  &  0.321  /  0.304  &  0.238  /  0.228  &  0.314  /  0.298 
\tabularnewline
yeast5 &  0.691  /  0.684  &  \textbf{0.738}  /  \textbf{0.730}  &  0.705  /  0.698  &  0.727  /  0.718 
\tabularnewline
yeast6 &  0.498  /  0.490  &  \textbf{0.575}  /  \textbf{0.567}  &  0.510  /  0.503  &  0.510  /  0.501 
\tabularnewline
\hline Average &  0.600  /  0.591  &  0.628  /  0.618  &  0.604  /  0.594  &  0.622  /  0.611 
\tabularnewline
\hline\end{tabular}\end{table*}

\begin{table*}[ht]\scriptsize\caption{Table showing F1-score/$\kappa$-score for different parameter settings for ConvGeN, using the GB classifier. The ConvGeN(min,maj) model is  seen to perform the best for all the tested classifiers.}\label{tab:results:GB:B}\centering\tabularnewline\begin{tabular}{l|@{\hskip3pt}c@{\hskip3pt}|@{\hskip3pt}c@{\hskip3pt}|@{\hskip3pt}c@{\hskip3pt}|@{\hskip3pt}c@{\hskip3pt}}\hline
\textbf{dataset ($f_1~$score$~/~\kappa~$score)} & \textbf{ConvGeN(5,maj)} & \textbf{ConvGeN(min,maj)} & \textbf{ConvGeN(5,prox)} & \textbf{ConvGeN(min,prox)}
\tabularnewline
\hline
abalone 17 vs 7 8 9 10 &  0.291  /  0.270  &  0.291  /  0.278  &  0.310  /  0.289  &  0.311  /  0.297 
\tabularnewline
abalone9 18 &  0.367  /  0.329  &  0.318  /  0.292  &  0.343  /  0.304  &  0.318  /  0.292 
\tabularnewline
car good &  0.714  /  0.705  &  0.864  /  0.858  &  0.733  /  0.723  &  \textbf{0.869}  /  \textbf{0.864} 
\tabularnewline
car vgood &  0.933  /  0.931  &  \textbf{0.982}  /  \textbf{0.982}  &  0.951  /  0.949  &  \textbf{0.982}  /  \textbf{0.982} 
\tabularnewline
flare F &  0.139  /  0.125  &  0.181  /  0.164  &  0.162  /  0.147  &  0.156  /  0.138 
\tabularnewline
hypothyroid &  0.789  /  0.777  &  0.800  /  0.789  &  0.789  /  0.777  &  \textbf{0.815}  /  \textbf{0.806} 
\tabularnewline
kddcup guess passwd vs satan &  0.995  /  0.995  &  \textbf{1.000}  /  \textbf{1.000}  &  0.996  /  0.996  &  \textbf{1.000}  /  \textbf{1.000} 
\tabularnewline
kr vs k three vs eleven &  \textbf{0.995}  /  \textbf{0.995}  &  \textbf{0.995}  /  \textbf{0.995}  &  0.992  /  0.992  &  \textbf{0.995}  /  \textbf{0.995} 
\tabularnewline
kr vs k zero one vs draw &  0.959  /  0.958  &  \textbf{0.972}  /  \textbf{0.971}  &  0.960  /  0.958  &  0.961  /  0.960 
\tabularnewline
shuttle 2 vs 5 &  \textbf{1.000}  /  \textbf{1.000}  &  \textbf{1.000}  /  \textbf{1.000}  &  0.994  /  0.994  &  \textbf{1.000}  /  \textbf{1.000} 
\tabularnewline
winequality red 4 &  \textbf{0.133}  /  \textbf{0.095}  &  0.107  /  0.066  &  \textbf{0.133}  /  \textbf{0.095}  &  0.102  /  0.063 
\tabularnewline
yeast4 &  0.257  /  0.242  &  0.351  /  0.329  &  0.261  /  0.245  &  0.333  /  0.311 
\tabularnewline
yeast5 &  0.713  /  0.705  &  0.724  /  0.716  &  0.731  /  0.723  &  0.714  /  0.705 
\tabularnewline
yeast6 &  0.490  /  0.481  &  \textbf{0.564}  /  \textbf{0.554}  &  0.481  /  0.472  &  0.520  /  0.510 
\tabularnewline
\hline Average &  0.627  /  0.615  &  0.654  /  \textbf{0.642}  &  0.631  /  0.619  &  0.648  /  0.637 
\tabularnewline
\hline\end{tabular}\end{table*}

\begin{table*}[ht]\scriptsize\caption{Table showing F1-score/$\kappa$-score for different parameter settings for ConvGeN, using the kNN classifier. The ConvGeN(min,maj) model is  seen to perform the best for all the tested classifiers.}\label{tab:results:KNN:B}\centering\tabularnewline\begin{tabular}{l|@{\hskip3pt}c@{\hskip3pt}|@{\hskip3pt}c@{\hskip3pt}|@{\hskip3pt}c@{\hskip3pt}|@{\hskip3pt}c@{\hskip3pt}}\hline
\textbf{dataset ($f_1~$score$~/~\kappa~$score)} & \textbf{ConvGeN(5,maj)} & \textbf{ConvGeN(min,maj)} & \textbf{ConvGeN(5,prox)} & \textbf{ConvGeN(min,prox)}
\tabularnewline
\hline
abalone 17 vs 7 8 9 10 &  0.290  /  0.261  &  0.315  /  0.296  &  0.273  /  0.243  &  0.311  /  0.291 
\tabularnewline
abalone9 18 &  0.361  /  0.305  &  0.136  /  0.120  &  0.347  /  0.290  &  0.165  /  0.149 
\tabularnewline
car good &  0.558  /  0.531  &  0.509  /  0.479  &  0.534  /  0.505  &  0.585  /  0.561 
\tabularnewline
car vgood &  0.659  /  0.641  &  \textbf{0.879}  /  \textbf{0.874}  &  0.700  /  0.684  &  0.794  /  0.784 
\tabularnewline
flare F &  0.269  /  0.220  &  \textbf{0.306}  /  \textbf{0.265}  &  0.277  /  0.228  &  0.302  /  0.262 
\tabularnewline
hypothyroid &  0.576  /  0.548  &  0.606  /  0.582  &  0.578  /  0.551  &  \textbf{0.646}  /  \textbf{0.628} 
\tabularnewline
kddcup guess passwd vs satan &  0.982  /  0.982  &  \textbf{0.991}  /  \textbf{0.991}  &  0.985  /  0.985  &  \textbf{0.991}  /  \textbf{0.991} 
\tabularnewline
kr vs k three vs eleven &  0.895  /  0.891  &  \textbf{0.961}  /  \textbf{0.960}  &  0.894  /  0.890  &  0.925  /  0.923 
\tabularnewline
kr vs k zero one vs draw &  0.864  /  0.858  &  \textbf{0.897}  /  \textbf{0.892}  &  0.865  /  0.859  &  0.853  /  0.847 
\tabularnewline
shuttle 2 vs 5 &  0.954  /  0.954  &  0.932  /  0.931  &  \textbf{0.960}  /  \textbf{0.960}  &  0.943  /  0.942 
\tabularnewline
winequality red 4 &  \textbf{0.074}  /  0.016  &  0.058  /  -0.001  &  \textbf{0.074}  /  0.016  &  0.059  /  0.000 
\tabularnewline
yeast4 &  0.334  /  0.296  &  0.348  /  0.313  &  0.338  /  0.301  &  0.345  /  0.309 
\tabularnewline
yeast5 &  0.650  /  0.636  &  0.646  /  0.631  &  0.640  /  0.624  &  0.659  /  0.645 
\tabularnewline
yeast6 &  0.337  /  0.311  &  0.426  /  0.406  &  0.334  /  0.308  &  0.390  /  0.367 
\tabularnewline
\hline Average &  0.557  /  0.532  &  0.572  /  0.553  &  0.557  /  0.532  &  0.569  /  0.550 
\tabularnewline
\hline\end{tabular}\end{table*}


\begin{table*}[ht]\scriptsize\caption{Table showing F1-score/$\kappa$-score for different parameter settings for ConvGeN, using the DoC classifier. The ConvGeN(min,maj) model is  seen to perform the best for all the tested classifiers.}\label{tab:results:DoG:A}\centering\tabularnewline\begin{tabular}{l|@{\hskip3pt}c@{\hskip3pt}|@{\hskip3pt}c@{\hskip3pt}|@{\hskip3pt}c@{\hskip3pt}|@{\hskip3pt}c@{\hskip3pt}}\hline
\textbf{dataset ($f_1~$score$~/~\kappa~$score)} & \textbf{ConvGeN(5,maj)} & \textbf{ConvGeN(min,maj)} & \textbf{ConvGeN(5,prox)} & \textbf{ConvGeN(min,prox)}
\tabularnewline
\hline
abalone 17 vs 7 8 9 10 &  0.306  /  0.278  &  \textbf{0.393}  /  \textbf{0.372}  &  0.317  /  0.289  &  0.366  /  0.345 
\tabularnewline
abalone9 18 &  0.477  /  0.433  &  0.538  /  0.512  &  0.512  /  0.473  &  \textbf{0.549}  /  \textbf{0.525} 
\tabularnewline
car good &  0.741  /  0.729  &  0.659  /  0.646  &  \textbf{0.759}  /  \textbf{0.748}  &  0.577  /  0.558 
\tabularnewline
car vgood &  0.772  /  0.763  &  0.721  /  0.712  &  \textbf{0.776}  /  \textbf{0.766}  &  0.753  /  0.743 
\tabularnewline
flare F &  \textbf{0.321}  /  \textbf{0.279}  &  0.223  /  0.193  &  0.302  /  0.262  &  0.267  /  0.234 
\tabularnewline
hypothyroid &  0.627  /  0.604  &  \textbf{0.703}  /  \textbf{0.688}  &  0.625  /  0.602  &  0.690  /  0.674 
\tabularnewline
kddcup guess passwd vs satan &  0.991  /  0.991  &  \textbf{1.000}  /  \textbf{1.000}  &  0.996  /  0.996  &  \textbf{1.000}  /  \textbf{1.000} 
\tabularnewline
kr vs k three vs eleven &  0.993  /  0.992  &  0.995  /  0.995  &  0.994  /  0.994  &  \textbf{0.996}  /  \textbf{0.996} 
\tabularnewline
kr vs k zero one vs draw &  0.957  /  0.955  &  0.951  /  0.949  &  \textbf{0.965}  /  \textbf{0.964}  &  0.941  /  0.939 
\tabularnewline
shuttle 2 vs 5 &  \textbf{1.000}  /  \textbf{1.000}  &  0.996  /  0.996  &  0.994  /  0.994  &  0.998  /  0.998 
\tabularnewline
winequality red 4 &  0.118  /  0.065  &  \textbf{0.123}  /  \textbf{0.073}  &  0.115  /  0.063  &  0.105  /  0.052 
\tabularnewline
yeast4 &  0.312  /  0.274  &  0.308  /  0.271  &  0.311  /  0.273  &  \textbf{0.334}  /  \textbf{0.299} 
\tabularnewline
yeast5 &  0.578  /  0.560  &  \textbf{0.652}  /  \textbf{0.639}  &  0.571  /  0.553  &  0.637  /  0.623 
\tabularnewline
yeast6 &  0.366  /  0.342  &  \textbf{0.491}  /  \textbf{0.475}  &  0.358  /  0.334  &  0.382  /  0.359 
\tabularnewline
\hline Average &  0.611  /  0.590  &  \textbf{0.625}  /  \textbf{0.609}  &  0.614  /  0.594  &  0.614  /  0.596 
\tabularnewline
\hline\end{tabular}\end{table*}

\par\mbox{~}\vspace{20cm}
\par\mbox{~}\vspace{20cm}
\par\mbox{~}\vspace{20cm}
\par\mbox{~}\vspace{20cm}
\par\mbox{~}\vspace{20cm}
\par\mbox{~}\vspace{20cm}
\par\mbox{~}\vspace{20cm}
\par\mbox{~}

\EOD

\end{document}